\documentclass{article} %
\usepackage{iclr2025_conference,times}

\usepackage{amsmath,amsfonts,bm}

\def\eqref#1{equation~\ref{#1}}

\def\1{\bm{1}}

\DeclareMathAlphabet{\mathsfit}{\encodingdefault}{\sfdefault}{m}{sl}
\SetMathAlphabet{\mathsfit}{bold}{\encodingdefault}{\sfdefault}{bx}{n}

\usepackage{url}
\usepackage[utf8]{inputenc} %
\usepackage[T1]{fontenc}    %
\usepackage[hypertexnames=false,colorlinks=true,citecolor=blue,backref=page]{hyperref}       %
\usepackage{booktabs}       %
\usepackage{amsfonts}       %
\usepackage{nicefrac}       %
\usepackage{microtype}      %
\usepackage{paralist}
\newcommand{\cmark}{\ding{51}}%
\newcommand{\xmark}{\ding{55}}%
\usepackage{makecell}
\usepackage{setspace}
\usepackage{multirow}
\usepackage{graphicx}
\usepackage{booktabs}
\usepackage{subcaption}
\usepackage{pifont}%
\usepackage{xspace}
\usepackage[font=small,tableposition=top]{caption}
\usepackage{float}
\floatstyle{plaintop}
\restylefloat{table}

\DeclareCaptionLabelFormat{andtable}{#1~#2  \&  \tablename~\thetable}

\newenvironment{lmtt}{\fontfamily{lmtt}\selectfont}{\par}

\newcommand{\todo}[1]{}
\newcommand{\new}[1]{#1}
\newcommand{\newrebuttal}[1]{#1}

\newcommand*\samethanks[1][\value{footnote}]{\footnotemark[#1]}
\newcommand{\eg}{\textit{e.g.}\xspace}

\newcommand{\OEmetric}{GEM}
\newcommand{\dsetsize}{3,268}
\newcommand{\dsetvideos}{2,405}
\def\sepappendix{0}

\title{Neptune: The Long Orbit to Benchmarking Long Video Understanding}

\author{Arsha Nagrani\thanks{Equal Contribution}, 
Mingda Zhang\samethanks, 
Ramin Mehran\samethanks, 
Rachel Hornung, \\
\textbf{Nitesh Bharadwaj Gundavarapu, 
Nilpa Jha, 
Austin Myers,
Xingyi Zhou,} \\ 
\textbf{Boqing Gong, 
Cordelia Schmid, 
Mikhail Sirotenko, 
Yukun Zhu, 
Tobias Weyand}\\
Google Research \thanks{Authors now at Google DeepMind}}

\iclrfinalcopy %
\begin{document}

\maketitle

\begin{abstract}
We introduce Neptune, a benchmark for long video understanding that requires reasoning over long time horizons and across different modalities. Many existing video datasets and models are focused on short clips (10s-30s). While some long video datasets do exist, they can often be solved by powerful image models applied per frame (and often to very few frames) in a video, and are usually manually annotated at high cost. In order to mitigate both these problems, we propose a scalable dataset creation pipeline which leverages large models (VLMs and LLMs), to automatically generate dense, time-aligned  video captions, as well as tough question-answer-decoy sets for video segments (up to 15 minutes in length). Our dataset Neptune covers a broad range of long video reasoning abilities and icludes a subset that emphasizes multimodal reasoning. Since existing metrics for open-ended question answering are either rule-based or may rely on proprietary models, we provide a new open source model-based metric (\OEmetric) to score open-ended responses on Neptune. \new{Benchmark evaluations reveal that most current open-source long video models perform poorly on Neptune, particularly on questions testing temporal ordering, counting and state changes. Through Neptune, we aim to spur the development of more advanced models capable of understanding long videos. The dataset is available at \href{https://github.com/google-deepmind/neptune}{https://github.com/google-deepmind/neptune}}.
\end{abstract}

\section{Introduction}
\label{sec:intro}
Videos are experiencing an \textit{explosion} moment online, with new research constantly pushing the frontier for video and language tasks such as video question answering (VideoQA)~\citep{xu2017video,zhong2022video,xiao2021next,yang2021just,mangalam2023egoschema}. Early video and language models, while adept at VideoQA, have largely focused on short, trimmed clips (less than 1 minute long~\citep{yu2019activitynet, xiao2021next}). The recent release of powerful, longer context multimodal models (eg. Gemini 1.5~\citep{reid2024gemini} and GPT4~\citep{achiam2023gpt}), however, has ushered in the promise of models being able to reason over millions of tokens, covering longer stretches of videos (many minutes long). 

While promising, these claims are often evidenced by qualitative examples, or results on small-size datasets -- for example the 1H-VideoQA~\citep{reid2024gemini} benchmark, which while valuable, only consists of 125 questions. Popular video benchmarks for question answering still tend to focus on short, trimmed clips (\eg, Next-QA~\citep{xiao2021next}). Other datasets that \textit{do} contain longer videos are often ‘short-term’ benchmarks disguised as long-term ones, evidenced by models that are able to solve them with a single (or a few) frames (\eg some tasks on the LVU dataset~\citep{wu2021towards} such as scene prediction of movies). Other long video datasets may contain strong linguistic biases in multiple choice evaluation, as shown by MoreVQA~\citep{min2024morevqa}, which gets strong performance on EgoSchema~\citep{mangalam2023egoschema} without access to the video at all, \new{or can be solved with external internet knowledge, such as those made from popular movies~\citep{li2023llama}.}

\begin{figure*}[t]
\centering
  \includegraphics[width=1.\linewidth]{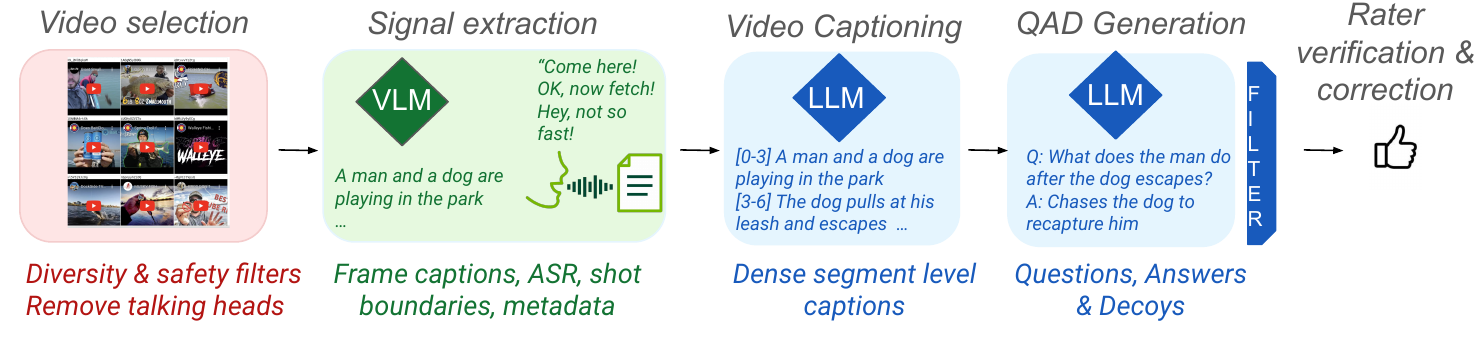}
  \caption{\textbf{Pipeline Overview:} Our pipeline consists of 5 key stages - (i) Video selection, where suitable videos are identified from YouTube, (ii) Signal extraction, (iii) Video level captioning, (iv) Question, answer and decoy (QAD) generation and (v) Manual rater verification. The first four stages are entirely automatic. Before rater verification, we automatically filter out QADs that can be solved by an LLM without access to the video content.
  }
  \label{fig:pipeline}
\end{figure*}

A key challenge in creating a truly long form video understanding dataset is the significant manual cost required to select, watch, understand and annotate long videos with free-form natural language. Answering challenging questions about longer videos is often a \textit{multimodal} (as it may involve listening to the audio track in addition to watching the video), and \textit{non-sequential}
endeavour (as sometimes it is necessary to rewind and rewatch key parts to answer a question). Proposing suitable high-level questions that are not trivially solved by a few frames is also tricky for humans to do consistently and with adequate diversity. The key aim of this paper is to solve this challenge by leveraging automatic tools to reduce rater effort while at the same retaining quality.
Inspired by EgoSchema, we do this by proposing a scalable dataset creation pipeline (Fig.~\ref{fig:pipeline}) that leverages strong foundational Video Language Models (VLMs) and Large Language Models (LLMs) with carefully designed prompts. We first generate dense, time-aligned video captions automatically, from which tough question-answer-decoy (QAD) sets can be automatically derived. This is done by extracting image captions, automatic speech recognition (ASR), shot boundaries and video metadata, and combining these signals with multi-stage, chain of thought prompting of an LLM. Our pipeline can be applied to any video on YouTube (Fig.~\ref{fig:pipeline}).

\new{While most of the pipeline is automatic, a comprehensive rater verification stage at the end ensures quality. While other dataset pipelines that are entirely manual~\citep{zhou2024mlvu,fang2024mmbench,wang2024lvbench}, our verification stage is lightweight, which we show by ablating the automatic part of the pipeline, and measuring the time taken by raters to propose QAs for videos from scratch. Results show that our semi-automatic pipeline almost halves rater effort.} Our dataset is called Neptune\footnote{Named after the planet with the longest orbit}, and covers a diverse range of videos, is multimodal (requires audio and visual information), and poses challenging questions for videos that test a variety of reasoning abilities over long time horizons.
Neptune allows for two modes of evaluation: multiple-choice and open-ended question answering. Since existing metrics for open-ended question answering are either rule-based and derived from captioning (WUPS~\citep{wu1994verbs}, CIDEr~\citep{vedantam2015cider}, etc) or are LLM-based evals that rely on proprietary APIs (such as ChatGPT\footnote{\url{https://openai.com/index/chatgpt/}}), we finetune the open source Gemma model~\citep{gemmateam2024gemma2} on a generic answer equivalence dataset~\citep{bulian2022tomayto} to score question answering results and evaluate it as a metric on a manually annotated answer equivalence dev set. We call this new metric Gemma Equivalence Metric (\OEmetric).

To summarise, we make the following contributions: 
(i) We propose a scalable pipeline to generate complex QAD annotations for any video that halves rater time compared to manual annotation.
(ii) We use this pipeline to generate the Neptune evaluation-only dataset, which consists of \new{\dsetsize{}} QAD annotations for \new{\dsetvideos{}} videos. We also release a \textit{challenging} subset, \textsc{Neptune-MMH} for which \textit{vision} plays an important role.
(iii) We provide both multiple choice and open-ended evaluation metrics. For the latter, we propose a new open-ended metric called Gemma Equivalence Metric (\OEmetric) which outperforms rule-based metrics on a manually annotated answer equivalence dataset; and finally 
(iv) We provide benchmarking and ablations of state-of-the-art VideoQA models on the Neptune sets. Benchmarking shows a significant gap between open-source video models and properietary models such as Gemini-1.5 and \new{GPT-4}.
All data will be released publicly to the research community.

\section{Related Works}
\label{sec:related_work}
\noindent\textbf{Video Question Answering:}
Video Question-Answering (VideoQA) is an important task for assessing multimodal video understanding systems' ability to reason about videos~\citep{xu2017video,zhong2022video,xiao2021next,yang2021just,mangalam2023egoschema}. Vision and language models for this task can be broadly classified into three categories: (i) early end-to-end VLMs for this task which typically consists of strong vision and language encoders/decoders, such as Flamingo~\citep{alayrac2022flamingo}, BLIP2~\citep{li2023blip2}, Video-Llama~\citep{damonlpsg2023videollama}, GIT2~\citep{wang2022git} and  PALI~\citep{chen2022pali,chen2023pali,chen2023pali3}. These typically are moderate sized models, and memory limits often lead to significant downsampling: \eg temporally sampling a few frames with large strides~\citep{wang2022git, chen2023pali} or spatially subsampling each frame to a single token~\citep{yang2023vid2seq, zhou2018end, wang2021end}; (ii) Socratic style models~\citep{zeng2022socratic}, which consists of combining various specialised \textit{frozen} models with carefully prompted state-of-the-art VLMs and LLMs (eg.  MoreVQA~\citep{min2024morevqa}) and (iii) end-to-end large multimodal models such as Gemini~\citep{team2023gemini} and GPT-4~\citep{achiam2023gpt}, which have long context lengths and can ingest multimodal data, including video, sound and text. \\
\noindent\textbf{Video QA Benchmarks:}
Key datasets have pushed towards assessing reasoning for temporal questions~\citep{grunde2021agqa,xiao2021next,wu2021star_situated_reasoning}, longer videos~\citep{yu2019activitynet,mangalam2023egoschema}, as well as focusing on diverse domains like instructional~\citep{yang2021just} and egocentric videos~\citep{Gao_2021_ICCV,mangalam2023egoschema}. We summarise existing VideoQA benchmarks in a table provided in the appendix. Most datasets either focus on shorter videos (less than 100s), or are short video datasets `in disguise', and can actually be solved with a few frames (\eg ActivityNet-QA~\citep{yu2019activityqa} or MovieQA~\citep{tapaswi2016movieqa}). 1H-VideoQA~\citep{reid2024gemini} consists of videos longer than 1 hour, but is limited to 125 questions and is closed-source. Like Neptune, ActivityNet-RTL~\citep{huang2024lita},
CinePile~\citep{rawal2024cinepile} and EgoSchema~\citep{mangalam2023egoschema} are generated by prompting LLMs, but cover only limited domains and rely on existing annotations while Neptune covers a much broader spectrum of video types and its pipeline is applicable to arbitrary videos. Most importantly,
 EgoSchema also has strong linguistic biases, while Neptune mitigates these through filtering (Sec. \ref{sec:experiments}).
Unlike other benchmarks which come with their own training sets (eg. MSR-VTT~\citep{xu2016msrvtt}, ActivityNet~\citep{yu2019activitynet}), we propose a generalisation-focused \textit{zero-shot} evaluation regime. The goal for Neptune is to benchmark any model, pre-trained with any external dataset or task, in order to assess real-world domain transfer. Hence we release \textit{test} sets only. More discussion on related datasets and dataset pipelines is provided in the appendix.

\noindent\textbf{Metrics for open-ended VideoQA:}
Earlier QA datasets consisted of short answers~\citep{xiao2021next} (sometimes a single word), typically from a closed set, and therefore metrics such as accuracy or accuracy with exact match (EM) can be applied. As datasets have evolved with more real-world annotation (longer, open-set answers), designing a metric becomes challenging. Existing rule-based metrics for captioning, such as BLEU~\citep{Papineni02}, ROUGE~\citep{lin2004rouge} and CIDEr~\citep{vedantam2015cider} can be applied, however they all primarily measure n-gram overlap, and do not capture the inherent subjectivity of the task, where different phrasing is often equally valid. Other metrics for captioning include SPICE~\citep{anderson2016spice} (adds action and
object relationships), while model-based metrics using earlier language models or image-language models include BERT-Score~\citep{zhang2019bertscore}, BERT-Score++~\citep{yi2020improving} (fine-tunes
BERT for image captioning),  LEIC~\citep{cui2018learning}, NUBIA~\citep{kane2020nubia}, TIGEr~\citep{jiang2019tiger}, CLIPScore~\citep{hessel2021clipscore}, \new{and EMScore~\citep{shi2022emscore}}. For answer equivalence specifically, token F1 and exact match (EM) have been used, but suffer many of the same shortcomings that rule-based metrics do, and EM is often too strict for open-ended eval.  BEM~\citep{bulian2022tomayto} finetunes BERT on an answer-equivalence dataset, and shows that this provides a better score for QA. Recently, LLMs trained with reinforcement learning from human feedback (RLHF) that already exhibit strong human alignment~\citep{bubeck2023sparks} are used in works such as VideoChatGPT~\citep{maaz2023video} and MovieChat~\citep{song2023moviechat} (LLM-as-a-judge). A challenge here is that the models used (ChatGPT) are called via proprietary APIs, where the underlying model may be non-static, thereby leading to non-reproducability in the metric. Instead, we take a state-of-the-art open-sourced lightweight language model~\citep{team2024gemma} and finetune it on a public answer equivalence dataset~\citep{bulian2022tomayto}, to create an open-source, static, model-based evaluation metric. 

\section{Neptune}
\begin{figure*}
\centering
  \includegraphics[width=1\linewidth]{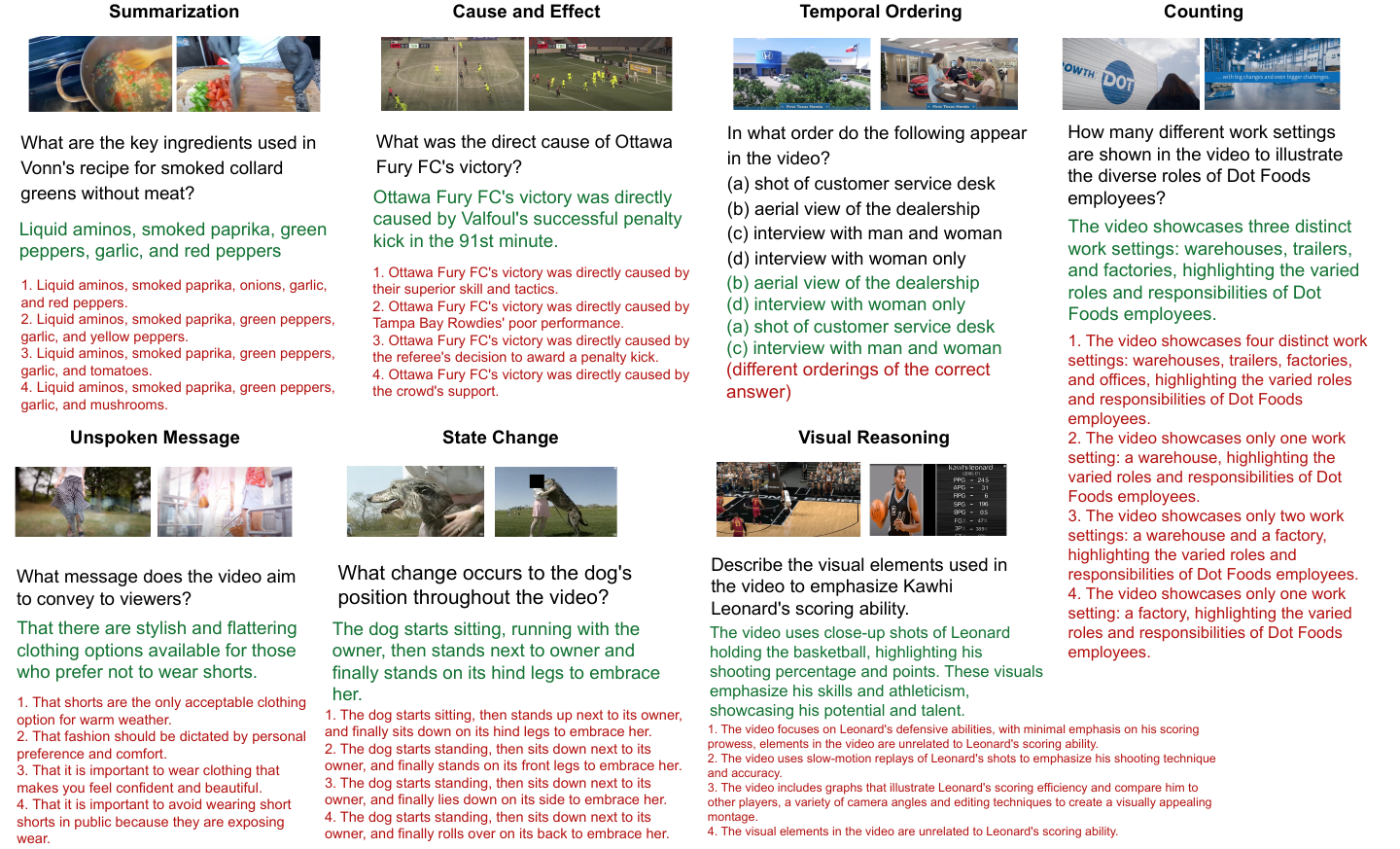}
  \caption{\textbf{Examples from Neptune:} We show examples from the dataset that highlight key question types from our dataset. We show 2 frames from each video. Correct answer is provided in green and decoys are shown in red. Best viewed zoomed in and in colour. Some decoys are summarised for brevity.
  }
  \label{fig:examples}
\end{figure*}

\begin{figure*}
  \begin{minipage}[b]{0.9\linewidth}
    \centering
    \includegraphics[width=\linewidth]{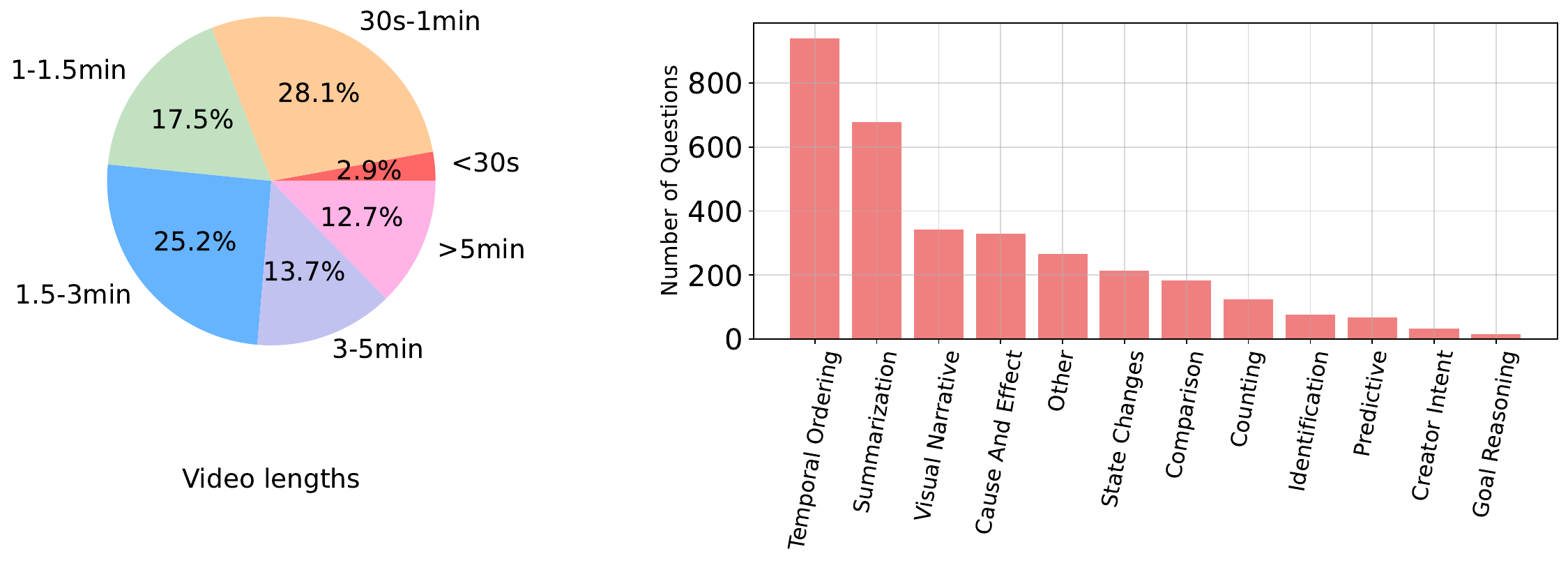}
  \end{minipage}%

    \includegraphics[width=1\linewidth]{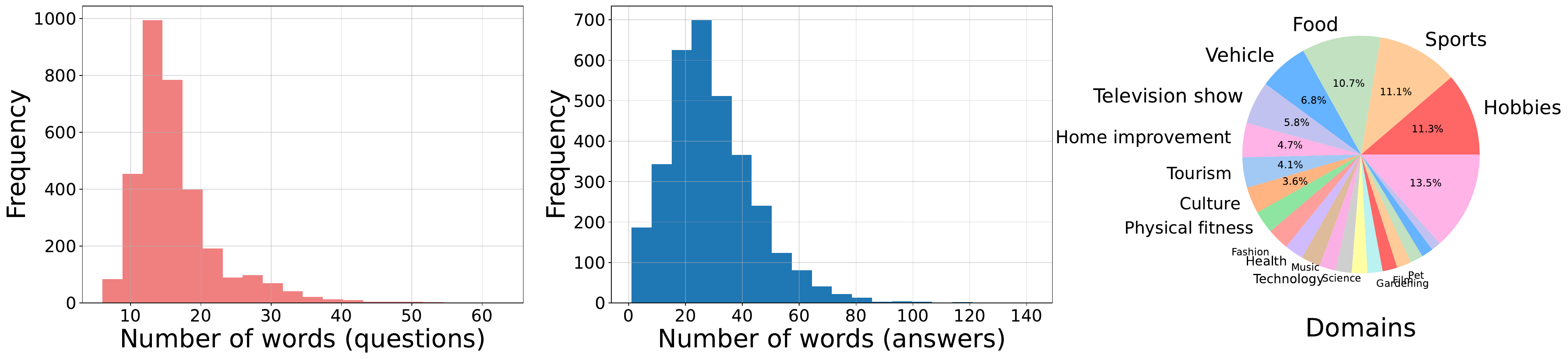}
\caption{\new{\textbf{Neptune Statistics:}} We show, the distribution of video lengths (top, left), the number of questions per question type (top, right), the distribution question and answer lengths (bottom, left and middle) and the domains in Neptune (bottom, right). Note that greater than 12\% of the videos are longer than 5 minutes (\new{305}) and over 25\% are longer than 3 minutes. An expanded plot of the video domains is provided in the appendix.}
\label{fig:statistics}
\end{figure*}
In this section we describe our dataset generated by the pipeline described in Sec.~\ref{sec:pipeline}. We first discuss motivating principles, which affect much of the prompt design in the pipeline stage (Sec. \ref{sec:pipeline}). Each video contains one or more annotation sets, which consists of a question, an answer to the question and four decoys (which are used for multiple choice evaluation).
Our key motivation is that questions should not be answerable by: 
(i) looking at a single (or few) frames; (ii) using text-only LLMs alone (language, common sense) that have no access to the video; (iii) 
with only the video’s speech transcript, and (iv) questions should cover a number of high-level `question types', which are discussed next and described in more detail in the appendix. \\
\noindent\textbf{Question Types.} 
Neptune covers a broad range of long video reasoning abilities, which are provided as `question type'
labels for each question. Examples are provided in Fig.~\ref{fig:examples}, and the distribution of questions per question type is depicted in Fig.~\ref{fig:statistics} (right). \new{More information about the distribution of question types is provided in the appendix}. Question types are obtained by carefully prompting an LLM (described in Sec.~\ref{sec:qa-generation}) and include \textit{Video Summarisation}, which involves summarising and comparing long parts of the video, as well as identifying the most important segments of the video; \textit{Visual Reasoning}, which involves understanding visual elements, as well as reasoning about why visual content is used (\eg to convey a certain mood); \textit{Temporal Ordering}, including the timeline of events; \textit{State Changes}; \textit{Counting} of higher level instances; \textit{Cause and Effect}, and understanding the \textit{Unspoken Message} or \textit{Creator Intent} in a video. \\
\noindent\textbf{Dataset Statistics.}
Our dataset consists of \new{\textbf{\dsetsize{}}} questions from \new{\textbf{\dsetvideos{}}} videos, covering \new{\textbf{100}} hours of video. We truncate videos longer than 15 minutes, with the smallest video being 16 seconds and the average length of videos being 2.5 minutes. We show the distribution of video lengths in Fig.~\ref{fig:statistics} (top, left). Note that greater than 12\% of the videos are longer than 5 minutes (\new{305} videos) and over 25\% are longer than 3 minutes, which is the maximum length of videos in the EgoSchema dataset. 
The distribution of questions per question type is depicted in Fig.~\ref{fig:statistics} (top, right). The most frequent question type is Temporal Ordering, followed by Summarization. Questions are on average \new{16.3} words long, while answers and decoys are \new{29.5} and \new{29.0} words long respectively. A full distribution of lengths can be seen in Fig.~\ref{fig:statistics} (bottom, left). We also note that the videos in Neptune cover a diverse range of topics (Fig.~\ref{fig:statistics} -- bottom, right), an expanded version of this plot is provided in the appendix.

\section{Dataset Creation Pipeline} \label{sec:pipeline}
An overview of our pipeline can be found in Fig.~\ref{fig:pipeline}. In order to reduce human effort, we leverage automatic tools to (i) find suitable videos (ii) extract useful signals and then (iii) automatically generate video level captions and QADs. We then send the data to human raters for the final manual verification stages. 
\new{Our pipeline can be applied to any generic YouTube video. This is unlike existing data pipelines such as those used to create EgoSchema~\citep{mangalam2023egoschema}, which relies on human generated captions, SFD~\citep{ghermi2024short} and other movie related datasets, which requires movie titles, loglines and synopses (human-written), or MLVU~\citep{zhou2024mlvu}, which re-uses annotations from existing datasets for many of their tasks. This makes the dataset scalable, as YouTube has a constantly growing set of videos.} Each stage is described in detail below. 
\vspace{-0.8em}
\subsection{Video Selection and Signal Extraction}
\vspace{-0.8em}
\label{sec:video-selection-signal-extraction}
\noindent\textbf{Video Selection:} We begin with the YT-Temporal-1Bn~\citep{zellers2022merlot} dataset. Because this dataset has strong speech and visual alignment, it consists of a lot of videos where \textit{`talking heads'} dominate the screen (eg. VLOGs, product placements, etc). We attempt to reduce the number of such videos in order to capture more interesting scenes, objects and actions. This is done by extracting face detections with frontal gaze where face bounding-box height is greater than 20\%, and removing videos where more than 30\% of frames have such frontal gaze. We then apply safety filters to remove racy, local controversy content etc, as well as applying filters to maximise semantic and person diversity. Details about these processes are provided in the appendix. \\
\noindent\textbf{Signal Extraction:} For each video we extract the following signals: (i) \textit{Frame captions:} A visual description of each frame (extracted at 1fps) is obtained from \new{PaLI-3 \citep{chen2023pali3}}.
(ii) \textit{ASR:} the speech is transcribed using the YouTube API; 
(iii) \textit{Metadata:} We obtain the YouTube title and the description for each video; and  
(iv) \textit{Shot boundaries} for each video. 
\vspace{-0.8em}
\subsection{Automatic Video Captioning}\label{sec:videocap}
\vspace{-0.8em}
The signals described above (frame captions, ASR, title and description, shot boundaries) are automatically combined to create video-level captions in a multi-stage process. \new{Examples of caption quality are provided in the appendix, showcasing details such as visual elements, multiple events, mood and atmosphere, details from the ASR, and even high level feelings and emotions.}
Video captions are obtained using the following steps: \\
\noindent\textbf{Shot Visual Captions}:
Using the shot boundaries, the \textit{frame captions} are summarized into shot-level descriptions (\textit{shot captions}) by prompting the same LLM. 
We then create a script for each video containing the shot timestamps, the shot visual captions and the ASR transcript. \\
\noindent\textbf{Topic and Description Pairs:} If ASR exists, an initial list of structured topics for the video (along with a short topic description) is formed by prompting an LLM with the ASR (see appendix). 
Note that this already yields decent topics as the initial list of videos have been selected (by the YT-Temporal-1Bn authors) to have a strong correlation between ASR and visual content. \\
\noindent\textbf{Shot Clustering:} Shots are then clustered per-video using an LLM prompted with the semantic topics obtained above. In each cluster, there may be one or many shots that correspond to that topic. A diagram on this stage and the exact prompt used is provided in the appendix. \\
\noindent\textbf{Segment Captions:} Consecutive shots of the same topic are then merged as one segment. Shots of the same topic that are not contiguous are treated as separate segments (see appendix for an example). We then generate dense captions for each segment using a custom prompt (see appendix). \\
\noindent\textbf{Adding Visual Support:}
To extract a better visual description of the segment that will be used for QA generation in the next phase, an extra step is performed to get visual support for each segment. That visual support is stored separately in conjunction with the dense caption for the segment. For this purpose, the dense caption from the previous step is used alongside the shot level visual captions. The LLM prompt used is provided in the appendix, and the the LLM used for all the above steps is Gemini-1.0-Pro \citep{team2023gemini}.

\subsection{QAD (Question-Answer-Decoy)  Generation} \label{sec:qa-generation}
\vspace{-0.8em}
We automatically generate questions, answers and decoys (QADs) by feeding the video captions from above to custom prompted LLMs. Our prompts are inspired by the EgoSchema dataset pipeline~\citep{mangalam2023egoschema}, with key modifications to generate more visually focused questions, as well as to generate questions belonging to a set of different question types. The exact prompts used are provided in the appendix. We generate QADs in two stages: (i) Given the video captions from the previous step, we first generate questions and answers; (ii) in the second stage we generate six decoys given the questions and answers from the previous stage. We found this 2-stage method to work better empirically than generating the QADs all in one go.

\subsection{LLM-based Blind Filter}\label{sec:blind-filter}
\vspace{-0.8em}
\noindent\textbf{QAD filter:} LLM-based generation can sometimes yield QAD triplets that can be answered from common sense or external world knowledge without the video as context. In particular, we observed that LLMs are often capable of inferring the correct answer from subtle cues in the answer candidates, for example if the correct answer is a positive sentiment while the decoys are negative. To remove such questions, we apply an LLM-based blind filter. We prompt an LLM (Gemini-1.0-pro) to rank the answer candidates to a question. To avoid false rejections due to random correct guesses, we repeat this process three times and only filter out questions where the model predicted the correct answer at least two times out of three (this number was selected to maximise number of videos left given the accuracy trade-off and is discussed in the appendix). Chain-of-thought reasoning improves accuracy so we ask the model to provide a rationale alongside its ranking.
\subsection{Manual Rater Verification} \label{sec:raters}
\vspace{-0.8em}
The final stage involves manual human verification. Raters are first asked to rate the quality of the question based on 4 criteria (details in the appendix). If the question is not suitable, the entire QAD set is discarded. If the question is accepted, raters annotate which modalities are required to answer the question. Choices are: ``audio+video'', ``video-only'', or ``audio-only''. Next, raters are asked to either accept the answer as-is or modify it. Decoys are annotated in a final stage. Given the six LLM-generated decoy candidates, raters verify that they are actually incorrect answers to the question and select the four most challenging ones. If less than four decoys are suitable, we provide a text field for raters to write their own decoys. After this step, a second rater verifies the QAD again to fix any errors the first rater might have missed. Screenshots of the rater UI are provided in the appendix. We noticed that rater corrections reintroduce a small amount of questions that can be answered without context, so as a final step we repeat the QAD filter described above. \new{We applied two rounds of manual rater verification to improve dataset quality and mitigate bias. So, overall each QAD has been verified by \emph{four} human raters. Overall, 65\% of QADs were rejected by raters. More details about rater training, replication (multiple raters per question) and pipelining are provided in the appendix.} \\
\noindent\textbf{Human Proposed Questions.}
\new{To test the effectiveness and efficiency of the automatic portion of our dataset pipeline, we asked raters to propose questions and answers entirely manually for a subset of the dataset. We call this set HPQ (Human Proposed Questions). The raters are provided with a few examples of each question type before they begin annotating. In total, we collect 270 QAs for 193 videos in this set. We use this set in two ways - (i) to quantitatively measure rater-time saved by our automatic pipeline, and (ii) to estimate the amount of Gemini bias in our semi-automatic pipeline. The results for both are provided in Sec.~\ref{sec:results}.}

\section{Experiments} \label{sec:experiments}
We now evaluate state of the art models, both open source and closed source, on Neptune and compare them to various baselines. We also analyze the effect of various parameters like the number of frames provided to the model. Before that, we introduce the two sets in Neptune and our evaluation metrics.
\vspace{-0.3cm}

\begin{figure}[t]
\begin{minipage}[t]{0.5\textwidth}%
              \centering
                \captionof{table}{Evaluation of open-ended metrics on \\ the \OEmetric~answer equivalence dev set. \\FT: Fine-tuning}
                \label{tab:gem-dev-results}
\resizebox{0.9\textwidth}{!}{\begin{tabular}{@{}lll@{}}
\toprule
\textbf{Metric} & \textbf{FT data} & \textbf{F1-Score}  \\

\midrule
CIDEr \citep{vedantam2015cider} & None & 56.4 \\
ROUGE-L \citep{lin2004rouge} & None & 62.2 \\
BEM \citep{bulian2022tomayto}  & BEM  & 61.5 \\
\midrule
Gemma-v1-2B-IT \citep{team2024gemma} &  None & 56.3\\
Gemma-v1-7B-IT  & None & 65.2 \\
Gemma-v2-9B-IT \citep{gemmateam2024gemma2} & None & 70.3 \\
Gemma-v2-9B-IT  (\OEmetric) & BEM  & \underline{71.2} \\
\midrule
Gemini-1.5-pro~\citep{reid2024gemini} & None & \textbf{72.8} \\
\bottomrule
\end{tabular}}
\end{minipage}%
\begin{minipage}[t]{0.5\textwidth}%
            \centering
            \captionof{table}{Results on the Human Proposed Question (HPQ) Split. *Results on \textsc{Neptune-Full} are reported on a subset containing the same set of videos as HPQ.}
            \label{tab:hpq}
                \resizebox{0.9\textwidth}{!}{\begin{tabular}{lrrrr}
                    \toprule
                   \textbf{ Method }& \textbf{Frames} & \textbf{ASR} & \textbf{\textsc{Full*}} & \textbf{\textsc{HPQ}} \\ 
                    \midrule 
                    Video-LLaMA-2 & 16 & No & 13.04 & 14.18 \\
                    MovieChat & 150 & No & 2.49 & 1.97 \\
                    MiniGPT4-Video & 45 & No & 5.14 & 4.10 \\
                    \midrule
                    Gemini-1.5-Pro & all & Yes & 45.05 & 44.44 \todo{comment out and replace with above results} \\
                    Gemini-1.5-Pro & all & No & 27.67 & 24.81 \\
                    \bottomrule

                \end{tabular}}
\end{minipage}%
\end{figure}

\subsection{Neptune Sets and Evaluation Metrics} \label{sec:sets}
\vspace{-0.8em}
\noindent\textbf{Neptune Sets:} Because we seeded our dataset from the YT-Temporal-1Bn~\citep{zellers2022merlotreserve} videos, it contains some videos where ASR can play a big role in contributing to the video content. In order to create a more challenging \textit{visual} benchmark, we also provide \textsc{Neptune-MMH} (multimodal human annotated), where we identify videos where vision should play an important role. This subset
is created by using the rater annotations for what modalities are required to answer the question (described in Sec.~\ref{sec:raters}), and discarding questions which the raters marked as solvable based on audio-only, \new{and consists of 1,171 QADs for 1,000 videos.}
We encourage the community to evaluate on this \textit{harder} subset as well. \\
\noindent\textbf{Evaluation:} We explore two different protocols for evaluation of question answering - multiple choice evaluation (which involves selecting the correct answer amidst 4 decoys), and open-ended evaluation, which involves producing an answer directly without any decoys and assessing answer quality directly. While the former has the advantage of easier metrics (simple accuracy), the latter removes any potential confounding biases in the decoys. Next, we outline our process for creating a new open-ended metric called \OEmetric. \\
\noindent\textbf{Gemma Equivalence Metric (\OEmetric):}
As discussed in Sec.~\ref{sec:related_work}, existing metrics for open-ended QA either lack robustness or rely on proprietary LLM APIs that can change over time. We therefore aim to produce a static open-ended metric. To evaluate how well different metrics align with human judgements, we construct a dev-set with \new{292} (question, reference answer, candidate answer, equivalence score) tuples. The equivalence scores were provided by humans and judge how well the candidate answer matches the reference answer on a scale between 0 and 1. \new{See Appendix~\ref{sec:appendix:gem} for details on the construction of the dev set.} We then benchmark a number of rule-based and model-based metrics on this set in Table~\ref{tab:gem-dev-results}.
To demonstrate the two ends of the scale, we first note that rule-based metrics such as CIDEr~\citep{vedantam2015cider} and ROUGE-L~\citep{lin2004rouge} obtain F1-Scores of 56.4 and 62.2, while an LLM-based metric using Gemini-1.5-pro~\citep{reid2024gemini} gets an F1-Score of \new{72.8} (but is closed-source). 
Next, we apply static open-source lightweight language models, namely the Gemma family of models i.e.\ \new{Gemma-v1-2B~\citep{team2024gemma},  Gemma-v1-7B~\citep{team2024gemma} and Gemma-v2-9B~\citep{gemmateam2024gemma2} to judge the answers in a zero-shot setting and find that performance improves with model size, with Gemma-v2-9B bridging the gap well between traditional metrics and the Gemini-1.5-pro based metric. Finally, we fine-tune Gemma-v2-9B on the open-source BEM answer equivalence dataset~\citep{bulian2022tomayto}. BEM contains answer equivalence ratings for questions about Wikipedia articles, so fine-tuning improves the model's ability to judge generic answer equivalence without making it specific to Neptune, or even to video. The resulting model performs the best on our dev-set among the Gemma models and closely approximates the performance of the much more expensive, closed-source Gemini-1.5-pro model.} We call the metric obtained with this model Gemma Equivalence Metric (\OEmetric{}). Note that this metric takes into account the question when comparing whether two answers are equivalent, which is unlike captioning metrics such as CIDEr which omit the question entirely.
 In Table \ref{tab:results_fixed}, we report open-ended evaluations using our proposed \OEmetric{} metric in addition to closed-ended MCQ accuracy. We will release \OEmetric{} publicly to enable reproducible open-ended evaluations. 
\vspace{-0.1in}

\subsection{Benchmarks} 
\vspace{-0.8em}
We describe all benchmarks used below. Implementation details are provided in the appendix. \\
\noindent\textbf{Blind Baselines:} \new{We evaluate models using a text-only prompt in two settings: (i) we feed only the question, answer and decoys to the model (QAD baseline).
(ii) we also feed ASR as an input for a QAD+ASR baseline. This helps identify questions that can be answered by prior or commonsense knowledge, or ASR only without obtaining visual information from video.}\\
\noindent\textbf{Image Models:} We use the BLIP2-T5-XL~\citep{li2023blip2} model, which contains a 1B vision encoder~\citep{EVA} and a 3B text-decoder~\citep{raffel2020exploring}. We feed the center frame of the video as the visual input, with prompt ``Answer in one letter'' followed by the question and shuffled answer and decoys. \new{We also evaluate some of the video models, \eg Gemini-1.5-pro and VideoLLaMA2, as image models, by feeding only the center frame.}\\
\noindent\textbf{Video Models:} We experiment with 3 different categories of VideoQA models: \\
\new{\noindent(i) Short Context MLLMs -  Video-LLaVA~\citep{lin2023video}, and VideoLLaMA2~\citep{cheng2024videollama}. We also experiment with a simple socratic JCEF (Just Caption Every Frame)~\citep{min2024morevqa}, which consists of a VLM to extract per-frame captions and an LLM to perform reasoning on top of these captions to answer the question. \\
(ii) Long Context MLLMs which are open-source, including MA-LMM~\citep{he2024ma}, MiniGPT4-Video~\citep{ataallah2024goldfish}, MovieChat~\citep{song2023moviechat},
LLaVA-OneVision~\citep{li2024llavaonevision}, InternVL2-8B~\citep{chen2024internvl15} and MiniCPM-v~\citep{yao2024minicpm}.
(iii) Long Context MLLMs which are closed-source, namely the Gemini 1.5 model family~\citep{reid2024gemini} and GPT-4o~\citep{achiam2023gpt}.} \\
\noindent\textbf{Implementation Details:}
For Video-LLaVA~\citep{lin2023video} we feed 8 uniformly sampled frames (resized to a minimum side length of 320 pixels) along with the question. We reimplement JCEF from the original paper~\citep{min2024morevqa} with updated components - i.e. $16$ uniformly sampled frame captions obtained using PaLI-3~\citep{chen2023pali}, and feed them as a text prompt to Gemini-1.0-pro along with the question and decoys. \new{For MiniGPT4-Video, we use the public codebase\footnote{https://github.com/Vision-CAIR/MiniGPT4-video} which routes videos longer than 3 minutes to their Goldfish model and those shorter to their older MiniGPT-video model.} 
We evaluate both the Gemini-1.5-pro and Gemini-1.5-flash models described in~\citep{reid2024gemini}.  We also experiment with feeding in ASR to the Gemini-1.5-pro model. For GPT-4o we use the public API\footnote{\label{footnote:gpt}Version gpt-4o-2024-05-13}. More details are provided in the appendix.\\

\begin{table}[t]
\centering
\resizebox{0.8\textwidth}{!}{
\begin{tabular}{@{}llrrrrr@{}}
\toprule
\textbf{Method}  & \textbf{ASR} & \textbf{Num. frames} & \multicolumn{2} {c} {\textsc{Neptune-Full}} & \multicolumn{2} {c} {\textsc{Neptune-MMH}} \\  & & & \textbf{Acc. \%} & \textbf{\OEmetric} & \textbf{Acc. \%} & \textbf{\OEmetric}  \\
\midrule
\textit{Open-source} \\
VideoLLaMA2~\citep{cheng2024videollama2}$\dagger$ & No & 0 & 38.31 & 4.91 & 30.03 & 0.88 \\ 
VideoLLaMA2~\citep{cheng2024videollama2} & Yes & 0 & \textbf{50.15} & \textbf{37.50} & 41.23 & \textbf{21.83} \\ 
VideoLLaMA2~\citep{cheng2024videollama2} & No & 1 (center) & 40.88 & 16.56 & 36.27 & 14.16 \\ 
VideoLLaMA2~\citep{cheng2024videollama2} & No & 4 & 43.92 & 16.87 & 39.61 & 10.62\\
VideoLLaMA2~\citep{cheng2024videollama2} & No & 8 & 44.74 & 16.26 & 41.32 & 15.93 \\
VideoLLaMA2~\citep{cheng2024videollama2} & No & 16 & 44.74 & 17.48 & 40.29 & 15.04 \\
VideoLLaMA2~\citep{cheng2024videollama2} & Yes & 16 & 49.28 & 32.54 & \textbf{45.38} & 18.18 \\
\midrule
\textit{Closed-source} \\
Gemini-1.5-pro~\citep{reid2024gemini}$\dagger$& No & 0 & 51.53 & 12.12  & 41.84 & 7.59 \\
Gemini-1.5-pro~\citep{reid2024gemini} & Yes & 0 & 76.68 & \textbf{44.92} & 65.76 & 31.20 \\
Gemini-1.5-pro~\citep{reid2024gemini} & No & 1 (center) & 55.57 & 14.11 & 51.75 & 13.27 \\
\new{Gemini-1.5-pro~\citep{reid2024gemini} & No & 8 & 63.80 & 18.74 & 59.95 & 18.87} \\
Gemini-1.5-pro~\citep{reid2024gemini} & No & 150 & 69.31 & 25.76 & 66.70 & 22.85 \\
Gemini-1.5-pro~\citep{reid2024gemini} & No & all & 68.94 & 25.40 &  65.58 & 23.44 \\
Gemini-1.5-pro~\citep{reid2024gemini} & Yes & all & \textbf{80.66} & \textbf{44.92}  & \textbf{75.32} & \textbf{34.87} \\
\bottomrule
\end{tabular} 
}
\caption{\new{\textbf{Ablations using different modalities and number of frames.}} $\dagger$ Blind baselines with no access to the video. We show results with one open-source and one closed-source video model.} \label{tab:mod_ablations} 
\end{table}

\begin{table}[h]
\centering
\resizebox{\textwidth}{!}{
\begin{tabular}{@{}llrrrr@{}}
\toprule
\textbf{Method}  & \textbf{Modalities} & \multicolumn{2} {c} {\textsc{Neptune-Full}}  & \multicolumn{2} {c} {\textsc{Neptune-MMH}} \\  &  & \textbf{Acc. \%} & \textbf{\OEmetric} & \textbf{Acc. \%} & \textbf{\OEmetric}  \\
\midrule
Random & - & 20.00 & & 20.00 & \\
\midrule
\textit{Image models} \\ 
BLIP2~\citep{li2023blip2} & RGB (center frame) & 34.80 & 9.20 & 28.10 & 8.50 \\
\midrule
\textit{Short Context MLLMs} \\
Video-LLaVA~\citep{lin2023video} & RGB (8 frames) & 25.79 & 10.66 & 24.00 & 5.48 \\
VideoLLaMA2~\citep{cheng2024videollama2} & RGB (16 frames) & 44.74 & 17.48 & 40.29 & 15.04\\ 
VideoLLaMA2~\citep{cheng2024videollama2} & RGB (16 frames) + ASR & 49.28 & 32.54 & 45.38 & 18.18 \\
\midrule
\textit{Long Context MLLMs - open-source} \\
MA-LMM~\citep{he2024ma} (ActivityNet-QA fine-tuned) & RGB (120 frames) & $\ddagger$20.22 & 10.67 & 19.51 & 5.04 \\ 
MiniGPT4-Video~\citep{ataallah2024goldfish} & RGB (45 frames) & 24.63 & 5.26 & 22.89 & 6.19 \\ 
MovieChat~\citep{song2023moviechat} & RGB (150 frames) & 28.96 & 3.79 & 30.30 & 1.01 \\
LLaVA-OneVision~\citep{li2024llavaonevision} & RGB (100 frames) & 66.22 & N/A & 62.82 & N/A\\
InternVL2-8B~\citep{chen2024internvl15} & RGB (16 frames) & 57.12 & 16.69 & 54.30 & 12.74 \\
MiniCPM-v~\citep{yao2024minicpm} & RGB (50 frames) & 56.59 & N/A & 53.27 & N/A\\
\midrule
\textit{Closed-source MLLMs} \\
VLM captions + LLM (JCEF)~\citep{min2024morevqa} & VLM captions (16 frames) & 58.51 & 12.27 & 56.45 & 11.50 \\
GPT-4o\textsuperscript{\ref{footnote:gpt}}~\citep{achiam2023gpt} & RGB (8 frames) + ASR & 80.23 & *49.01 & 72.86 & N/A\todo{add} \\
\new{Gemini-1.5-pro-002~\citep{reid2024gemini} & RGB (all frames) + ASR & \textbf{80.98} & \textbf{49.83} & \textbf{75.42} & \textbf{37.08}} \\
\todo{Optional: } Gemini-1.5-flash~\citep{reid2024gemini} \todo{Replace with Gemini-1.5-Flash-002 results} & RGB (all frames) + ASR & 76.90 & 45.59 & 71.05 & 33.93 \\
\bottomrule
\end{tabular} 
}
\caption{\textbf{Benchmarking performance on Neptune.} \textit{All frames:} Visual frames extracted at 1fps. *Computed on 10\% of the results. $\ddagger$ MCQ performance is close to random.}
\label{tab:results_fixed} 
\end{table}

\begin{figure}
\centering
\begin{minipage}{.6\textwidth}
  \centering
  \includegraphics[width=1\linewidth]{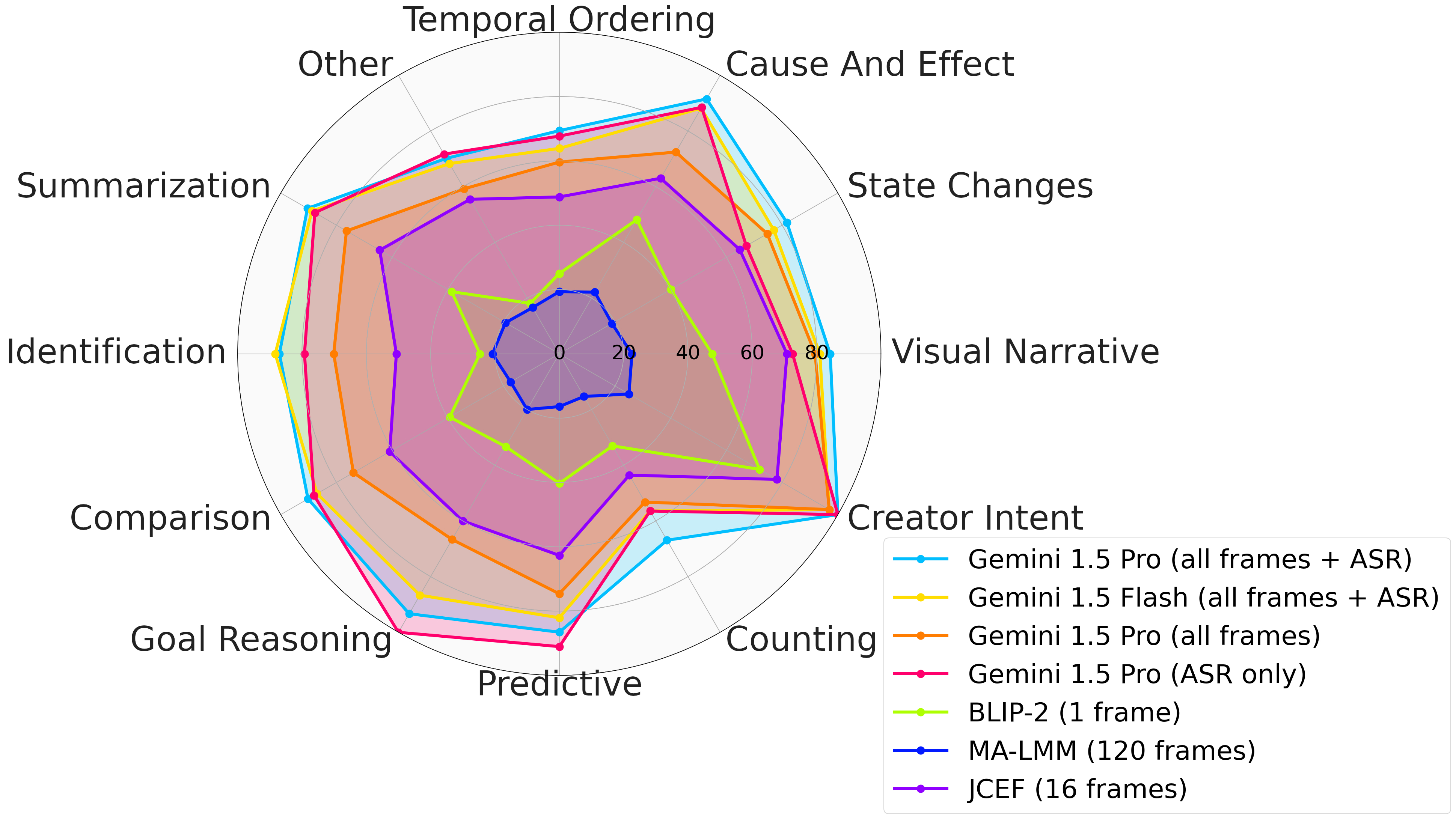}
\end{minipage}%
\begin{minipage}{.4\textwidth}
  \centering
  \includegraphics[width=1\linewidth]{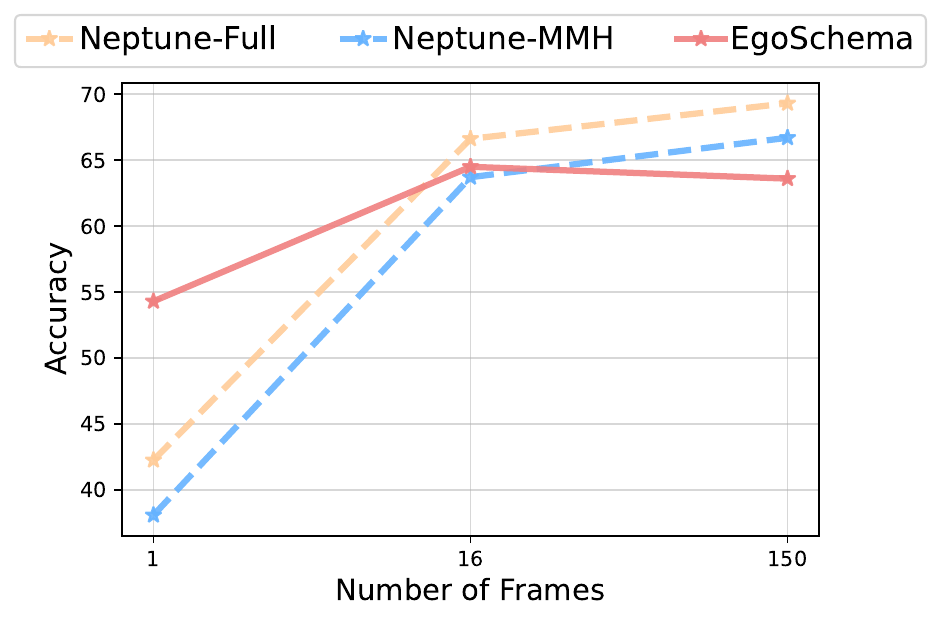}
\end{minipage}
\caption{\textbf{Performance of different models across question types on \textsc{Neptune-Full} (left) and Neptune Vs Egoschema with different frame rates (right).} On the right we show Gemini 1.5 Pro’s accuracy
when linearly subsampling to 1, 16 or 150 frames. We note that (i) performance on the Neptune sets increases as more frames are provided while on EgoSchema it saturates after 16 frames and (ii) \textsc{Neptune-MMH} is more challenging than EgoSchema. We included additional comparisons to other datasets in the appendix in Sec.~\ref{sec:appendix:dataset_comparison}.}
\label{fig:results-breakdown}
\vspace{-0.3cm}
\end{figure}
\vspace{-0.3cm}
\subsection{Results} \label{sec:results}
Results for all the baselines applied to the two Neptune sets (Sec. \ref{sec:sets}) are provided in Table \ref{tab:results_fixed}. We provide blind baselines and modality ablations in Table \ref{tab:mod_ablations} for VideoLLaMA2 and Gemini-1.5-pro.
\\
\noindent\textbf{Single-frame baselines and frame ablation:} We examine performance of the BLIP2 image-only model (Tab.~\ref{tab:results_fixed}) and two video models (VideoLLaMA2 and Gemini-1.5-pro) with only the center frame of the video in Tab.~\ref{tab:mod_ablations}. The larger Gemini model outperforms BLIP-2, however single-frame performance is much lower than with multiple frames. Performance on the middle frame of the video is generally better than on the first frame. For example, Gemini-1.5-pro achieves 55.57\% on the middle frame and 42.26\% on the first frame (see Fig.~\ref{fig:results-breakdown}, right). This is likely because the first frame is often blank or contains a title or intro. VideoLLaMA2's performance saturates at 8 frames because it is a short context model while Gemini's performance keeps increasing until 150 frames. We provide additional frame ablations in the appendix in Sec.~\ref{sec:appendix:frame_ablation}.\\
\noindent\textbf{Modality Ablations:} Table \ref{tab:mod_ablations} shows that performance of Gemini-1.5-pro and VideoLLaMA2 with ASR only as input is higher than performance with multiple video frames on the \textsc{Neptune-Full} set, but the gap shrinks significantly on the \textsc{Neptune-MMH} set, especially for MCQ eval. Surprisingly, the best result of VideoLLaMA2 is obtained using ASR only and not providing image frames. In fact, if we provide 16 frames in addition to ASR (last row of the open-source block), performance drops slightly. This may be a result of attention dilution \citep{coleman2023incontextinterference}, where an increasingly large context distracts the model, causing a drop in performance. For Gemini-1.5-pro on both sets however, the best performance is obtained with both frames and ASR, showcasing the complementary nature of the modalities.

\noindent\textbf{Video Models:} \new{Interestingly, we find that some open-source models that are designed specially for longer context video understanding (MA-LMM~\citep{he2024ma}, MiniGPT4-Video~\citep{ataallah2024goldfish} and MovieChat~\citep{song2023moviechat}) perform worse than VideoLLaMA2. This was also found by concurrent datasets such as MLVU~\citep{zhou2024mlvu} and LVBench~\citep{wang2024lvbench}.
The gap between many open-source and proprietary large MLLMs is also shown on concurrent datasets, \eg LVBench~\citep{wang2024lvbench}, where MovieChat gets near-random results and Gemini-1.5-pro is the state-of-the-art. One reason for this near-random performance may be the domain gap between the training sets of these models~\citep{he2024ma,song2023moviechat} and Neptune -- MovieChat is trained on movies and MA-LMM is designed to be fine-tuned on downstream QA datasets. More recent open source models (LLaVA-OneVision~\citep{li2024llavaonevision}, InternVL2-8B~\citep{chen2024internvl15} and MiniCPM-v~\citep{yao2024minicpm}) have made significant advances. Notably, LLaVA-OneVision reaches close to the performance of Gemini without ASR. By not providing a training set, we intentionally aim to assess generalization via zero-shot performance. We also note that the simple JCEF baseline, which consists of frame captions fed to an LLM for reasoning, outperforms most open-source models. The gap performance gap between open-source and closed-source MLLMs suggests Neptune may be a challenging benchmark for the future development of open-source models for long videos.}\\
\noindent\textbf{Challenging split and Gemini Bias:}  
 GPT-4o and Gemini-1.5-pro perform comparably on \textsc{Neptune-full}, despite Gemini-1.5-pro being used in dataset creation. On \textsc{Neptune-MMH}, neither model is able to achieve saturated performance. This suggests that Neptune does not exhibit significant bias towards Gemini models, even though Gemini was used in its construction. We believe that our extensive human verification helped mitigate bias. This is unlike VideoVista~\citep{li2024videovista} which uses GPT-4 to generate QADs automatically. However the performance of GPT-4 and Gemini-1.5 on their dataset is close to saturated (98\% on some categories). We note that performance falls for all models universally on the \textsc{Neptune-MMH} set, demonstrating the challenging nature of this set.\\
\noindent\textbf{Results on HPQ and model bias:} \new{In Tab.~\ref{tab:hpq}, we compare open-ended question answering performance on questions generated by our pipeline to performance on fully human written questions (HPQ) on the same set of videos. The time taken to manually create HPQ (19.03 minutes on average per question) is significantly longer than simply discarding or correcting QAs generated automatically as is done in our Neptune pipeline (10.32 minutes). While most models perform slightly worse on HPQ, overall performance is similar, suggesting that our automatic pipeline reaches the same difficulty level with roughly half the rater effort. Notably, Gemini-1.5-pro performs comparatively on both sets, suggesting that bias introduced by Gemini and PaLI-3 in the creation pipeline is limited.}\\
\noindent\textbf{Video Coverage compared to EgoSchema:}
Fig.~\ref{fig:results-breakdown} graphs the performance of Gemini-1.5-pro as a function of the number of frames provided on the two Neptune sets and Egoschema. For 1 frame, we take the first frame of the video. Gemini-1.5-pro’s performance on Neptune increases until 150 frames, while on EgoSchema it saturates after 16 frames, suggesting that Neptune's questions require seeing more frames to answer. Note that every video in EgoSchema has 180 frames (3 mins), whereas Neptune has variable lengths, with videos up to 15 minutes long. Results with the first frame on both Neptune splits are also much lower than those on EgoSchema (54.3\%), pointing to higher single-frame bias in the latter. To measure how much of a video a model needs to see to answer a given question, EgoSchema introduced the concept of a temporal certificate. We introduce a slightly modified version, which is \textit{model-based}, and show that the Gemini-1.5-pro model needs more frames to answer a question correctly in Neptune, with a mean certificate of 5.39 frames (compared to 1.6 for EgoSchema). The details of this experiment are provided in the appendix in Sec.\ref{sec:appendix-temporal-certificate}. We also provide comparisons to other long video benchmarks in the appendix in Sec.~\ref{sec:appendix:dataset_comparison}. \\
\noindent\textbf{Open-ended results:} We find that in general, results with \OEmetric{} mirror the trends demonstrated by the multiple choice eval, with the exception of the Gemini-1.5-flash and Gemini-1.5-pro results, as well as the performance of \new{the long context open-source models}. Here we find that Gemini-1.5-flash actually slightly exceeds the  performance of Gemini-1.5-pro on the \textsc{Full} set, and \new{MovieChat performs worse on the open-ended task than other baselines, while better on the MCQ evaluation}. A qualitative examination of the scores with the highest disparity shows that Gemini-1.5-flash seems to indeed provide better open-ended answers. Examples of this are provided in the appendix in Sec.~\ref{sec:appendix:gem-examples}.\\
\noindent\textbf{Results per question type:} Performance of different models across the different question types are shown in Fig. \ref{fig:results-breakdown}. We find that “Counting”, “Temporal Ordering” and “State Change” questions are challenging for all models, pointing to areas for future work for video-language models, while “Cause and Effect” is easier. Interestingly, the Gemini-1.5-Pro model applied only to ASR without access to video frames is the best at “Goal Reasoning”, which may be because human goals in videos are often mentioned in speech. Yet as expected, it is worse at the “Visual Narrative” questions, where Gemini-1.5-Pro models with access to RGB frames do much better.

\section{Conclusion}
\vspace{-0.1in} 
\label{sec:conclusion}
We present Neptune, a new benchmark for VideoQA with a focus on \emph{multimodal}, \emph{high-level} understanding of \emph{long videos}. Neptune is created using a scalable pipeline for arbitrary videos that minimizes (though not omits) human verification. Benchmarks are evaluated using MCQ and open-ended evals -- for which we provide a new, open-source metric. \noindent\textbf{Limitations:} The dataset may inherit biases of the Gemini model used to generate QADs. 
While VideoQA is a good proxy for video understanding, our dataset could be further improved by additional annotations -- such as manually annotated temporal grounding, dense captions or entity labels.

\bibliography{neptune1}
\bibliographystyle{iclr2025_conference}

\pagebreak
\section{Appendix}

\if\sepappendix1{\tableofcontents}\else{}\fi

\appendix
\if\sepappendix1{
\section{Errata}
\begin{table}[h]
\centering
\resizebox{\textwidth}{!}{
\begin{tabular}{@{}llllllll@{}}
\toprule
\textbf{Method}  & \textbf{Modalities} & \multicolumn{2} {c} {\textsc{Neptune-Full}} & \multicolumn{2} {c} {\textsc{Neptune-MMA}} & \multicolumn{2} {c} {\textsc{Neptune-MMH}} \\  &  & \textbf{Acc. \%} & \textbf{\OEmetric} & \textbf{Acc. \%} & \textbf{\OEmetric} & \textbf{Acc. \%} & \textbf{\OEmetric}  \\

\midrule
Random & - & 20.00 & & 20.00 & & 20.00 & \\
\midrule
BLIP2~\citep{li2023blip2} & RGB (first frame) & 36.23 & 43.28 & 23.79 & 41.86 & 27.81 & 40.86 \\
Gemini-1.5-pro~\citep{reid2024gemini} & RGB (first frame) & 41.63 & 5.73** & 35.11 & 5.10** & 38.39 & 6.04** \\
\midrule
PaLI-3~\citep{chen2023pali} & RGB (8 frames) & 20.30$\ddagger$ & 20.08 & $\ddagger$ & 17.49 & $\ddagger$ & 18.63\\
Video-LLaVA~\citep{lin2023video} & RGB (8 frames) & 25.58 & 40.88 & 19.38 & 39.64 & 21.57 & 36.19 \\
JCEF~\citep{min2024morevqa} & VLM captions (16 frames) & 45.37 & 32.57 & 29.64 & 30.55 & 38.87 & 32.83\\
\midrule
Gemini-1.0-pro$\dagger$ & QAD only & 32.01$\pm$0.09 & 35.56 & N/A* & 32.02 & N/A* & 31.18 \\
Gemini-1.0-pro$\dagger$ & QAD+ASR only & 59.03$\pm$0.33 & 59.28 & N/A* & 50.37 & 47.14$\pm$0.80 & 44.57 \\
\midrule 
Gemini-1.5-flash~\citep{reid2024gemini} & RGB (all frames) & 63.25 & 51.30 & 49.15 & 52.67 & 60.10 & 49.31\\
Gemini-1.5-pro~\citep{reid2024gemini} & RGB (all frames) & 68.95 & 50.09 & 55.28 & 49.26 & 64.70 & 47.60 \\
Gemini-1.5-pro~\citep{reid2024gemini} & RGB (150) + ASR & 79.69 & 63.77 & 63.12 & 59.24 & 71.91 & 56.32 \\
\bottomrule
\end{tabular} 
}
\caption{\textbf{Benchmarking performance on Neptune.} \textbf{All frames:} Visual frames extracted at 1fps. $\dagger$ Blind baselines with no access to the video. $\ddagger$ MCQ performance is close to random. This is discussed in the text. *Cannot be applied as this model was used to create this set. **Majority open-ended responses are variations of "I cannot answer due to limited information."} \label{tab:results_fixed} 

\end{table}
We note that in Table 2, the multiple-choice QA results of \textsc{Neptune-MMA} should be switched with \textsc{Neptune-MMH}. We include a corrected version of the table in Tab~\ref{tab:results_fixed}.
}\else{}\fi

\section{Related Works} 
Here we provide an additional discussion of related works that were omitted from the main paper due to lack of space.  
The recently released Perception Test~\citep{patraucean2024perception} consists of script-based recorded videos with manual annotations focusing on 4 broad skill areas - Memory, Abstraction, Physics, Semantics, however videos are only 23s long (avg).
Like Neptune, ActivityNet-RTL~\citep{huang2024lita} was constructed in a semi-automatic fashion by querying GPT-4 to generate comparative temporal localization questions from the captions in ActivityNet-Captions~\citep{krishna2017anetcaption}. %
CinePile~\citep{rawal2024cinepile} was generated by prompting an LLM to generate multiple-choice questions. Because it is based on movie clips, it can leverage available human-generated audio descriptions. Both ActivityNet-RTL and CinePile cover only limited domains and rely on existing annotations while Neptune covers a much broader spectrum of video types and its pipeline is applicable to arbitrary videos. 
Our rater stage is lightweight, unlike other works that are entirely manual~\citep{zhou2024mlvu,fang2024mmbench,wang2024lvbench}. In LVBench~\citep{wang2024lvbench}, even the video selection is done manually, and for MoVQA~\citep{zhang2023movqa}, only the decoys are generated automatically. Another recently released dataset (concurrent with our submission) is the Video-MME dataset~\citep{fu2024video}. The motivation of this dataset is similar to ours, namely it covers videos of variable lengths, with 2,700 QADs covering a wide range of different question types. The main difference between Video-MME and Neptune is that the former is entirely manually annotated by the authors, while we propose a scalable pipeline which can be applied to new videos and domains automatically, and can be tweaked to include different question types with reduced manual effort. EgoSchema is the closest work to ours in motivation, but there are some key differences: (i) it is limited to egocentric videos of exactly 3 minutes each, while Neptune covers many domains and follows a more natural length distribution for online videos (16s to 15min); (ii) it relies heavily on manually obtained dense captions for egocentric videos, while our method generates captions automatically too and hence can be easily applied to any video online; and more importantly (iii)
 EgoSchema also has strong image and linguistic biases, while Neptune mitigates these.
\begin{table}[h]
\resizebox{\textwidth}{!}{
\begin{tabular}{@{}llcrrrc@{}}
\toprule
\textbf{Name}  & \textbf{Ann} 
& \textbf{Rater V} & \textbf{Avg.\ len (s)} & \textbf{\# Vids (total/test)} & \textbf{\# Samples (total/test)} & \textbf{Available} \\ \midrule
MovieQA~\citep{tapaswi2016movieqa}        & QAD          
& \cmark         & 200              & 6,771/1,288                     & 6,462/1,258                        & \xmark $\dagger$ \\
MSRVTT-QA~\citep{xu2017video}      & QA           
& \xmark         & 15               & 10,000/2,990                    & 243,680/72,821                     & \cmark          \\
ActivityNet-QA~\citep{yu2019activitynet} & QA           
& \cmark         & 180              & 5,800/1,800                     & 58,000/18,000                      & \cmark           \\
NExTQA~\citep{xiao2021next}         & QAD          
& \cmark        & 44               & 5,440/1,000                     & 52,044/8,564                       & \cmark           \\
IntentQA~\citep{li2023intentqa} & QAD 
& \cmark &44&4,303/430 &16,297/2,134&\cmark\\ 
EgoSchema~\citep{mangalam2023egoschema}      & QAD         
&     \cmark   &    180 &    5,063/5,063 & 5,063/5,063 & \cmark$\ddagger$ \\

Perception Test~\citep{patraucean2024perception} &QAD
& \cmark & 23 & 11,600 & 38,000 &\cmark\\ 
MVBench~\citep{li2023mvbench} & QAD  & \xmark & 16 &  3,641 &  4,000 & \cmark \\ 
Video-Bench~\citep{ning2023video} & QAD & \cmark & 56 & 5,917 & 17,036 & \cmark \\ 
AutoEval-Video~\citep{chen2023autoeval} & QA & \cmark & 14.6 &  327 & 327 & \cmark \\ 
1H-VideoQA~\citep{reid2024gemini} & QAD & \cmark & 6,300 (max) & 125 & 125 & \xmark  \\
MLVU~\citep{zhou2024mlvu} & QAD & \cmark & 720 & 2K & 2593 & \cmark \\

\newrebuttal{Video-MME~\cite{fu2024video}} & \newrebuttal{QAD} & \newrebuttal{\cmark} & \newrebuttal{82.5/562.7/2,385.5**} & \newrebuttal{900} & \newrebuttal{2,700} & \newrebuttal{\cmark} \\ 
\newrebuttal{LongVideoBench~\cite{wu2024longvideobench}} & \newrebuttal{QAD} & \newrebuttal{\cmark}& \newrebuttal{473} & \newrebuttal{3,763} & \newrebuttal{6,678} & \newrebuttal{\cmark} \\
\midrule
\textbf{Neptune} & QAD 
& \cmark & \new{150/901*} & \new{\dsetvideos{}} & \new{\dsetsize{}} & \cmark \\
\textbf{Neptune-MMH} & QAD 
& \cmark & 159/901* & 1,000 & 1,171 & \cmark \\
\bottomrule
\end{tabular}
}
\caption{\textbf{Comparison to Existing VideoQA datasets:} \textbf{Ann. Type:} Annotation Type, \textbf{QAD:} Question, Answer and Decoys, \textbf{Rater V:} Rater verified manually. $\dagger$ Movies are no longer available. $\ddagger$ Annotations are hidden behind a test server, 500 are public. *average/max length. **short/medium/long.} \label{tab:qadatasets}
\end{table}
\section{The Neptune Dataset}
\subsection{Additional Information on Question Types} 
Neptune covers a broad range of long video reasoning abilities, which are summarised below. These question types are obtained in the Question and Answer generation stage, for which the prompt is provided in Sec. \ref{sec:prompt-QAgen}. We provide further insights into the motivations of some of the question areas provided in the prompt below. \\
\noindent\textbf{Video Summarisation:} Summarise and compare long parts of the video, as well as identify the most important segments of the video. \\
\noindent\textbf{Visual Reasoning:} Recognize and understand visual elements in different parts of the video, as well as reason about why visual content is used (\eg to convey a certain mood).\\
\noindent\textbf{Temporal Ordering:} Understand the timeline of events and the plot in the video.\\
\noindent\textbf{Counting:} Count objects, actions and events. Here we focus on higher-level counting where the same instance does not occur in all/every frame and actions are sufficiently dissimilar.\\
\noindent\textbf{Cause and Effect:} Understand and reason about cause and effect in the video.\\
\noindent\textbf{Message:} Understand the unspoken message that the audience may perceive after watching the video, which may require common sense knowledge to infer.\\
\noindent\textbf{State Changes:} Understand object states change over time, such as a door opening and food being eaten. \\

Since the questions are proposed automatically by an LLM, the question types are also generated in an open-set manner by the LLM. Hence sometimes, the LLM will generate the question type label using different phrasing - eg. `temporal ordering' or `timeline event'. We use simple manual postprocessing to group similar question types into the same category, with a few question types that do not fall into any of the categories grouped as `Other'. The final \textit{question types} released with the dataset are shown in Fig. \ref{fig:statistics} of the main paper.
\subsubsection{Question Type Distribution}
\begin{figure}[t]
\includegraphics[width=\linewidth]{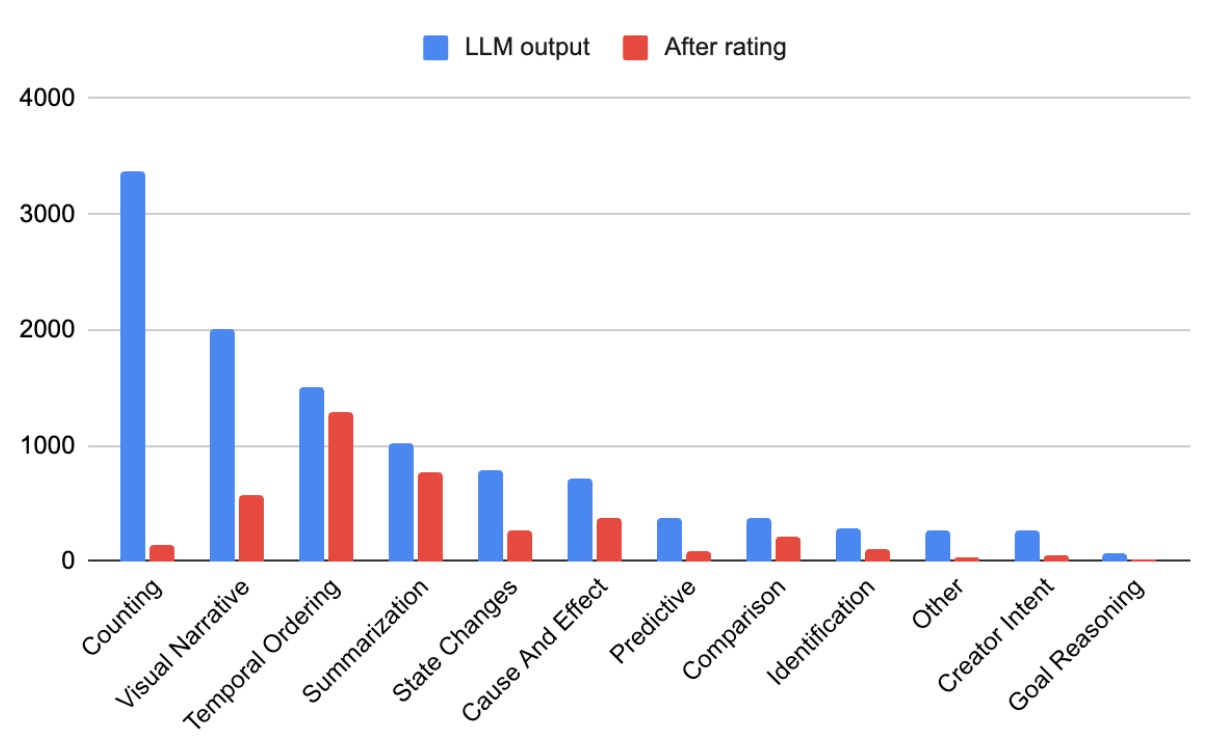}
\caption{\newrebuttal{Change of question type distribution as a result of human rater filtering.}}
\label{fig:question_types_change}
\end{figure}
We explain the reasons for Neptune’s current question type distribution: \\
(i) We prompted the LLM that generated the questions with a set of examples of different question types and let the model choose which questions to generate. \\
(ii) The model’s selection of question types depends strongly on the given video. For example, while it is always possible to ask for a video summary, it is not always possible to ask about a person’s goals, or cause and effect, because not all videos allow for these types of reasoning. This naturally leads to an imbalance of possible question types. \\
(iii) Additionally, we observed that the quality of questions produced by the LLM varies strongly by question type. Therefore, after quality checking by raters, the distribution changes significantly \newrebuttal{(Fig.~\ref{fig:question_types_change})}. The strongest difference was for counting questions, as LLM-proposed questions were often too easy, e.g. counting the number of times a certain word is mentioned.

\subsection{Domains in Neptune} 
\begin{figure*}[t]
\centering
  \includegraphics[width=1.\linewidth]{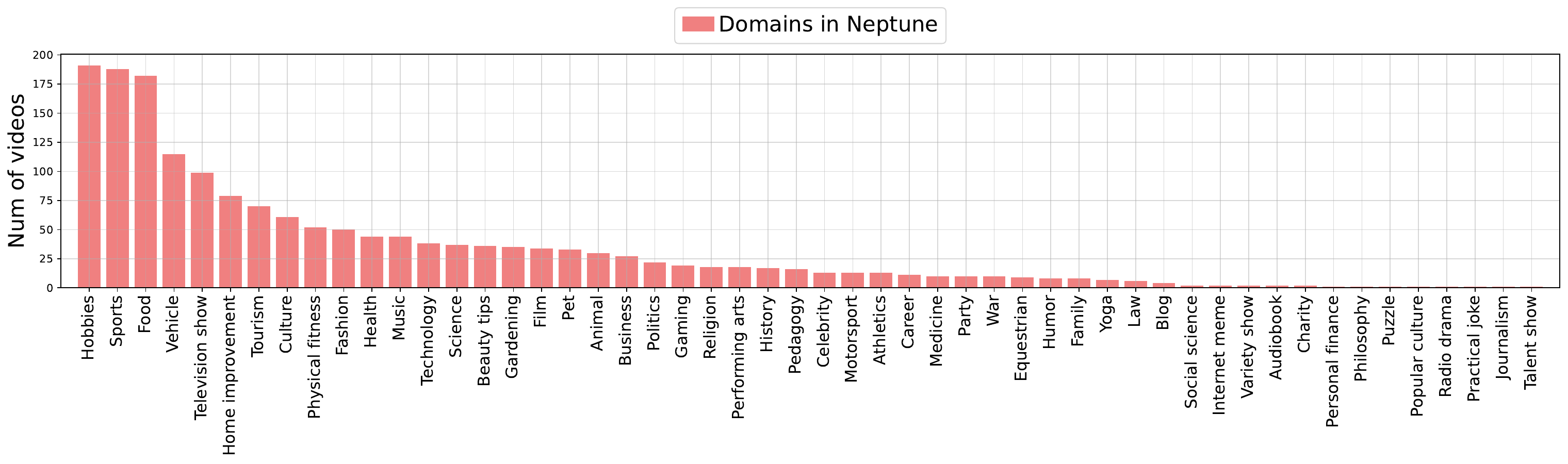}
  \caption{\textbf{Domains in Neptune:} We show the number of videos per domain category in \textsc{Neptune-full}. 
  }
  \label{fig:domains-extended}
\end{figure*}

A full graph of the domains in Neptune are provided in Fig. \ref{fig:domains-extended}.

\newrebuttal{\subsection{Comparison to other benchmarks}}
\label{sec:appendix:dataset_comparison}
\begin{figure}[t]
    \centering
    \includegraphics[width=\linewidth]{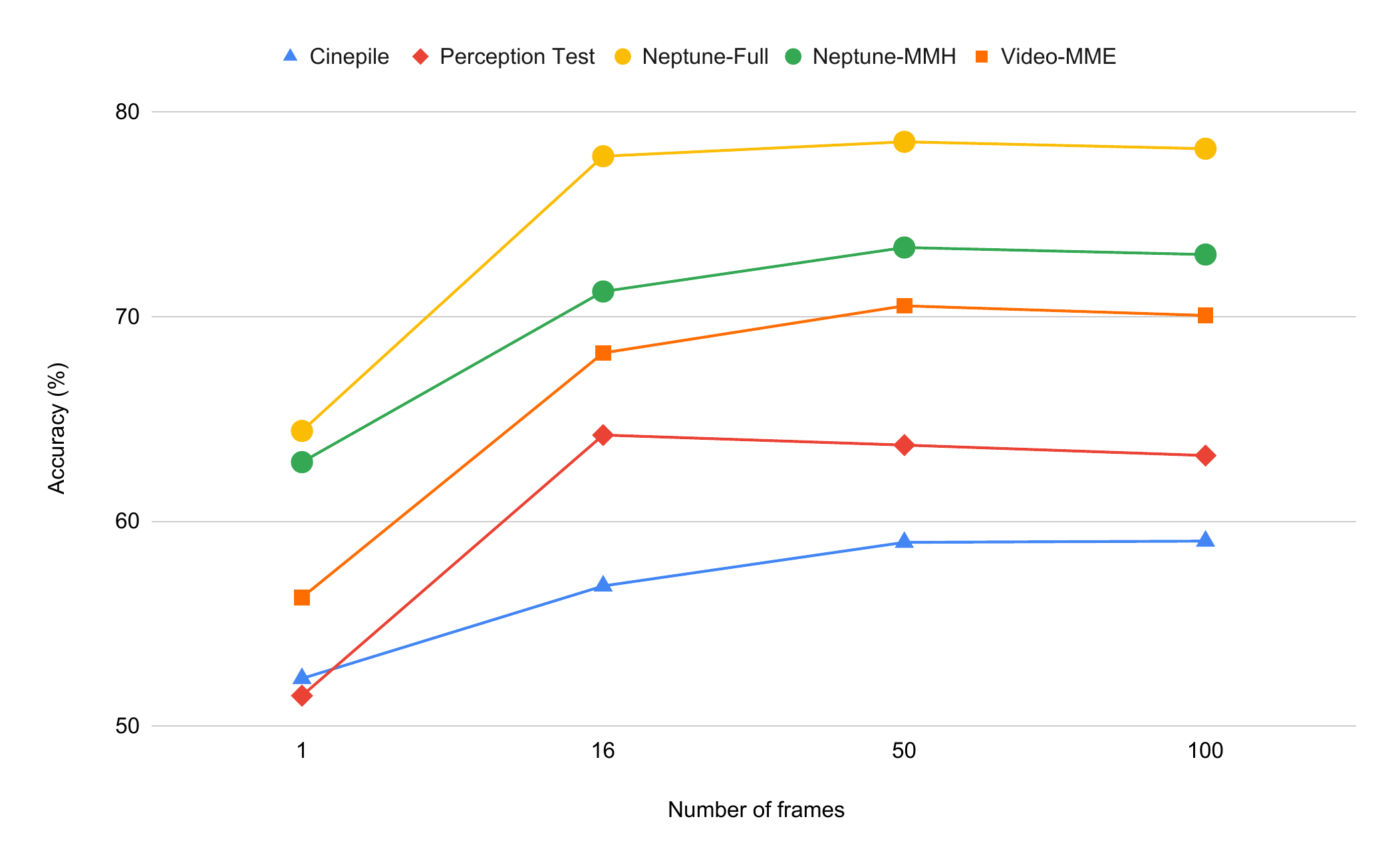}
    \caption{\newrebuttal{Comparison of Neptune to other video benchmarks. We evaluate Gemini-1.5-Flash with different numbers of frames that are uniformly sampled across the videos.}}
    \label{fig:dataset_comparison}
\end{figure}
\newrebuttal{We measure the complexity of Neptune compared to other benchmarks by analyzing the progression of model performance as we add more frames to the context. We use Gemini-1.5-Flash for this comparison since it is capable of handling very large contexts. Fig.~\ref{fig:dataset_comparison} shows the results of this experiment, comparing Neptune with CinePile~\citep{rawal2024cinepile}, Perception Test~\citep{patraucean2024perception} and Video-MME~\citep{fu2024video}. We find that most benchmarks saturate at about 50 frames, including Video-MME, which has much longer videos than Neptune. While we included Perception Test here as it is new, it does not claim to be a long video benchmark and saturates at 16 frames.}
\subsection{\newrebuttal{Per-Task performance}}
\begin{table}[t]
    \centering
    \resizebox{\textwidth}{!}{
    \begin{tabular}{llrrrrrrrrrrrrr}
        \toprule
        \textbf{Method} & \textbf{Modalities} & \textbf{TO} & \textbf{CE} & \textbf{SC} & \textbf{VN} & \textbf{CI} & \textbf{CT} & \textbf{PR} & \textbf{GR} & \textbf{CMP} & \textbf{ID} & \textbf{SUM} & \textbf{OTH} & \textbf{Task-avg} \\
        \cmidrule(lr){1-2} \cmidrule(lr){3-14}  \cmidrule(lr){15-15}
         \textit{Image models} \\ 
BLIP2~\citep{li2023blip2} & RGB (center frame) & 24.97 & 48.18 & 40.09 & 47.51 & 71.88 & 33.06 & 40.30 & 33.33 & 39.34 & 24.68 & 38.64 & 18.11 & 38.34 \\
\midrule
\textit{Short Context MLLMs} \\
Video-LLaVA~\citep{lin2023video} & RGB (8 frames) & 22.95 & 36.06 & 28.30 & 30.79 & 46.88 & 19.35 & 35.82 & 53.33 & 31.15 & 20.78 & 20.06 & 23.40 & 30.74 \\
VideoLLaMA2~\citep{cheng2024videollama2} & RGB (16 frames) & 35.71 & 57.27 & 48.36 & 57.31 & 78.13 & 33.06 & 53.73 & 60.00 & 50.54 & 42.86 & 47.35 & 29.81 & 49.51 \\ 
VideoLLaMA2~\citep{cheng2024videollama2} & RGB (16 frames) + ASR & 34.08 & 60.00 & 53.77 & 59.06 & 90.63 & 38.71 & 56.72 & 73.33 & 61.96 & 54.55 & 59.00 & 35.09 & 56.41 \\
\midrule
\textit{Long Context MLLMs - open-source} \\
MA-LMM~\citep{he2024ma} & RGB (120 frames) & 19.34 & 22.12 & 18.87 & 22.58 & 25.00 & 15.32 & 16.42 & 20.00 & 17.49 & 20.78 & 19.32 & 16.60 & 19.49 \\
MiniGPT4-Video~\citep{ataallah2024goldfish} & RGB (45 frames) & 20.43 & 34.24 & 24.06 & 30.79 & 34.38 & 21.77 & 31.34 & 33.33 & 21.31 & 32.47 & 23.16 & 21.89 & 27.43 \\ 
LLaVA-OneVision~\citep{li2024llavaonevision} & RGB (100 frames) & 57.83 & 73.33 & 72.99 & 77.71 & 84.38 & 41.94 & 82.09 & \underline{86.67} & 67.21 & 71.43 & 71.39 & 55.47 & 70.20 \\
MiniCPM-V 2.6~\citep{yao2024minicpm} & RGB (50 frames) & 41.32 & 65.15 & 67.3 & 70.38 & 75.0 & 37.9 & 67.16 & \underline{86.67} & 60.66 & 66.23 & 66.22 & 46.42 & 62.53 \\
\midrule
\textit{Closed-source MLLMs} \\
JCEF~\citep{min2024morevqa} & VLM captions (16 frames) & 48.78&	63.03&	64.79&	70.76&	78.13&	43.55&	62.69&	60.00&	60.87	&50.65&	64.45&	55.47&	60.26 \\
GPT-4o\textsuperscript{\ref{footnote:gpt}}~\citep{achiam2023gpt} & RGB (8 frames) + ASR & \textbf{71.25} & \underline{91.21} & \underline{77.25} & 76.83 & \textbf{100.0} & \underline{62.90} & \textbf{89.55} & \textbf{93.33} & \underline{87.98} & 85.71 & \textbf{91.30} & \textbf{72.45} & \underline{83.31} \\
Gemini-1.5-pro~\citep{reid2024gemini} & RGB (all frames) + ASR & \underline{69.39} & \textbf{91.52} & \textbf{81.69} & \textbf{84.21} & \textbf{100.0} & \textbf{66.94} & \underline{86.57} & \textbf{93.33} & \textbf{90.22} & \underline{87.01} & \underline{90.41} & \underline{70.19} & \textbf{84.29} \\
Gemini-1.5-flash~\citep{reid2024gemini} & RGB (all frames) + ASR & 63.87 & 88.18 & 77.00 & \underline{80.99} & \underline{96.88} & 56.45 & 82.09 & \underline{86.67} & 86.96 & \textbf{88.31} & 88.79 & 68.30 & 80.37 \\
\bottomrule
    \end{tabular}
    }
    \caption{\newrebuttal{Per-task model performance. Tasks are abbreviated as follows: TO: Temporal Ordering, CE: Cause And Effect, SC: State Changes, VN: Visual Narrative, CI: Creator Intent, CT: Counting, PR: Predictive, GR: Goal Reasoning, CMP: Comparison, ID: Identification, SUM: Summarization, OTH: Other. The best accuracy per task is printed in bold and the second best underlined.}}
    \label{tab:per-task-performance}
\end{table}
\newrebuttal{We provide detailed per-task model performance in Tab.~\ref{tab:per-task-performance}. See Fig.~\ref{fig:results-breakdown} (bottom left) for a graphical representation of a subset of these results. Overall, closed-source MLLMs perform best across all tasks, with Gemini-1.5-pro ranking best overall and GPT-4o ranking second. Even though their average scores are close, there are significant differences in per-task scores, showing the different capabilities of each model.}

\subsection{Frame Ablation}
\label{sec:appendix:frame_ablation}
\begin{figure}[t]
\includegraphics[width=\linewidth]{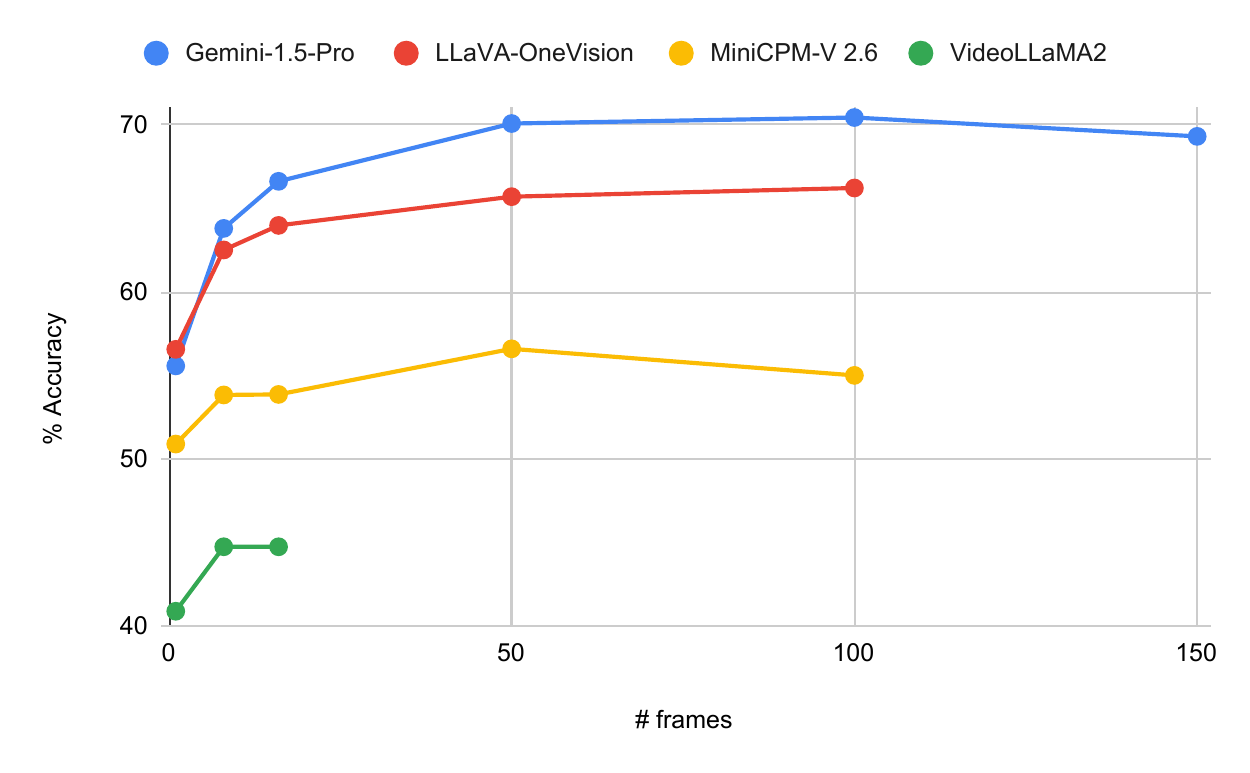}
\caption{Performance of different VLMs when providing 1, 8, 16, 50, 100 and 150 frames sampled uniformly across the video.}
\label{fig:frame_ablation}
\end{figure}
We compare the performance of Gemini-1.5-Pro~\citep{reid2024gemini} and three open-source models when providing varying numbers of frames. Frames are sampled uniformly across the video and we do not provide ASR. Results are show in Fig.~\ref{fig:frame_ablation}. We find that Gemini saturates at around 50 frames while open-sourced models have varying saturation points: VideoLLaMA2~\citep{cheng2024videollama2} saturated at 8 frames, MiniCPM~\citep{yao2024minicpm} at 50 frames and LLaVA-OneVision~\citep{li2024llavaonevision} at 100 frames. For both MiniCPM and LLaVA-OneVision we were unable to fit more frames into the context window.

\section{Implementation Details}
\subsection{Video Selection}
We choose the YT-Temporal-1Bn dataset \citep{zellers2022merlotreserve} as the source for Neptune, because of its large and diverse corpus, and because of the high correlation between vision and audio transcripts.

\textbf{Safety \& Content Filters:} We filter out videos with less than $100$ views, that are uploaded within $90$ days, and those tagged by YouTube content filters to contain racy, mature or locally controversial content. We then identify and remove static videos (eg.\ those that consist of a single frame with a voiceover) by clustering similar frames in a video and ensure that there is more than $1$ cluster. We also identify and remove videos comprising primarily of "talking heads". To achieve this, we apply a per-frame frontal-gazing face-detector at 1fps and mark the frames where the bounding box height is greater than $20\%$ as \textit{talking head frames}. Then, we filter out videos where more than $30\%$ of the frames are talking head frames. These thresholds are chosen based on an F1-score on a small dev set of $50$ manually annotated videos.

\begin{figure}[ht]
\centering
  \includegraphics[width=0.5\linewidth]{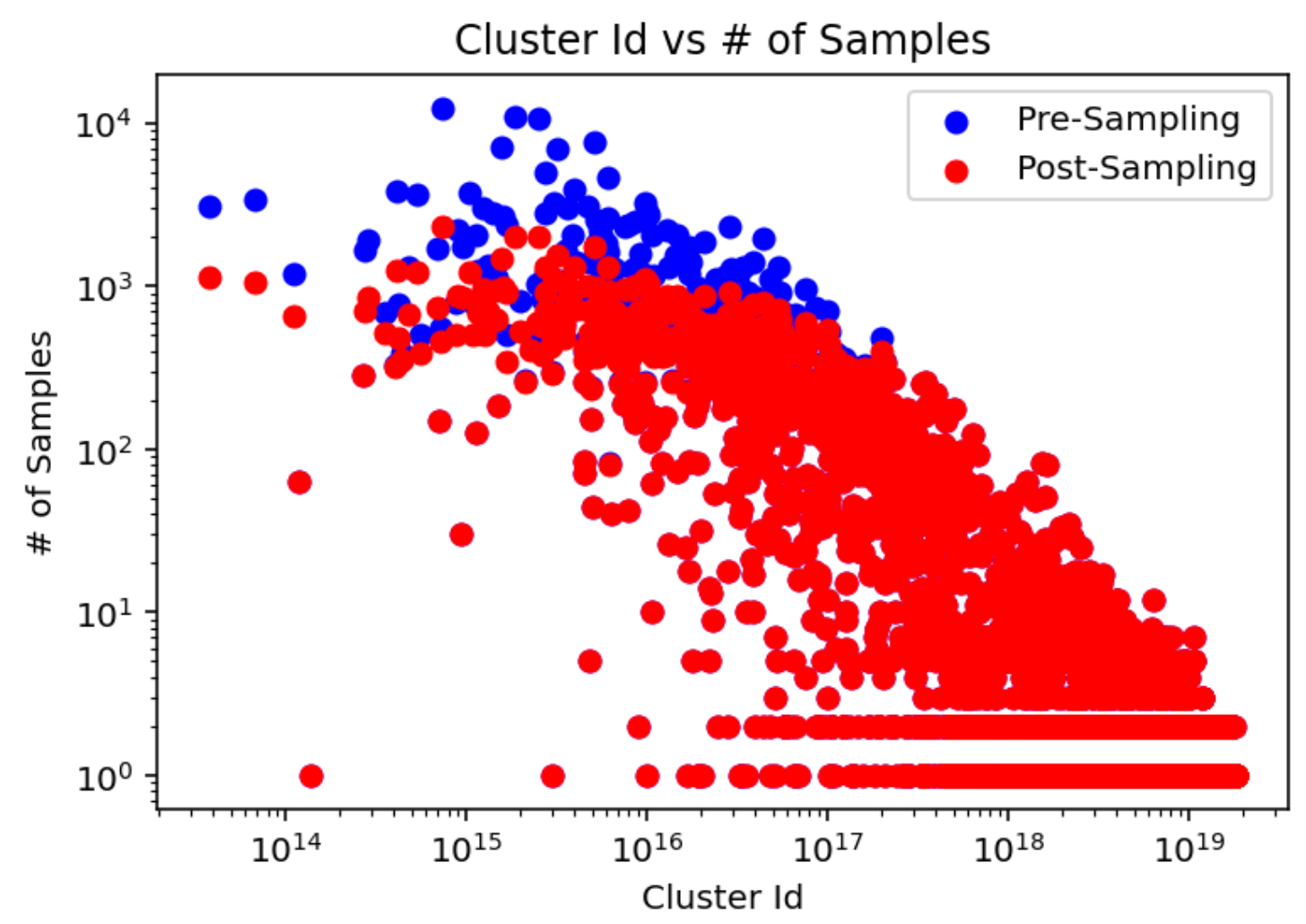}
  \caption{\textbf{Diversity sampling:} We show the change in cluster distribution after diversity sampling. 
  }
  \label{fig:diversity-sampling}
\end{figure}

\textbf{Diversity Sampling:} From the filtered set of videos, we sub-sample $100,000$ videos to boost both semantic and demographic diversity. First, we cluster the videos based on video-level semantic embeddings and tag each video with a cluster id. Second, we tag each video with the perceived age and gender demographic information contained in the video. Third, we obtain a joint distribution of semantics (cluster id) and demographics (perceived age and gender) and apply a diversity boost function \citep{kim2022improving} on the joint distribution. Finally, we sample from videos from this distribution. Fig.~\ref{fig:diversity-sampling}, shows the down-sampling of over-represented cluster ids before and after applying the filter.
We then uniformly sub-sample the videos further to reach the desired dataset size.

\subsection{Prompts for Data Generation}\label{supp:prompts}
In this section we provide some of the prompts used for generating Neptune. 
\subsubsection{Prompt for Frame Captioning}
We use the following prompt to obtain a caption for each video frame:

\begin{lmtt}
 Answer the following questions about the given image. Then use the information from the answers only, and write a single sentence as caption. Make sure you do not hallucinate information.

 Question(Mood): Describe the general mood in the image as succinctly as possible. Avoid specifying detailed objects, colors or text.
  \newline
 Question(Background): Describe the background of the image as succinctly as possible. Avoid specifying detailed objects, colors or text. Eg: The background is a parking lot, playground, kitchen etc.
  \newline
 Question(Person): Is there any person in the image. If yes, describe them and what are they doing here? If no, say no person.
  \newline
 Question(General): Describe the image as succinctly as possible. Avoid specifying detailed objects, colors or text.
 \newline
 Question(Text): Is there any text? What does it say?

 Result template:

 Answer(Mood): A succinct description of what is happening in the image with the general mood.
 \newline
 Answer(Background): A succinct description of the background scene in the image and what is happening. 
 \newline
 Answer(Person): If there are people in the image, a succinct description. 
 \newline
 Answer(General): A succinct description of the image.
  \newline
 Answer(Text): Reply if there is any text, where it is placed and how it is related to what is happening in the image.

 Caption: A couple of sentences summarizing the information given by the answers about mood, background, person, general and text.

 With the above format as template, generate the response for the new image next.
\end{lmtt}

\subsubsection{Prompts for Automatic Video Captioning}
\begin{figure}
    \centering
    \includegraphics[width=1\linewidth]{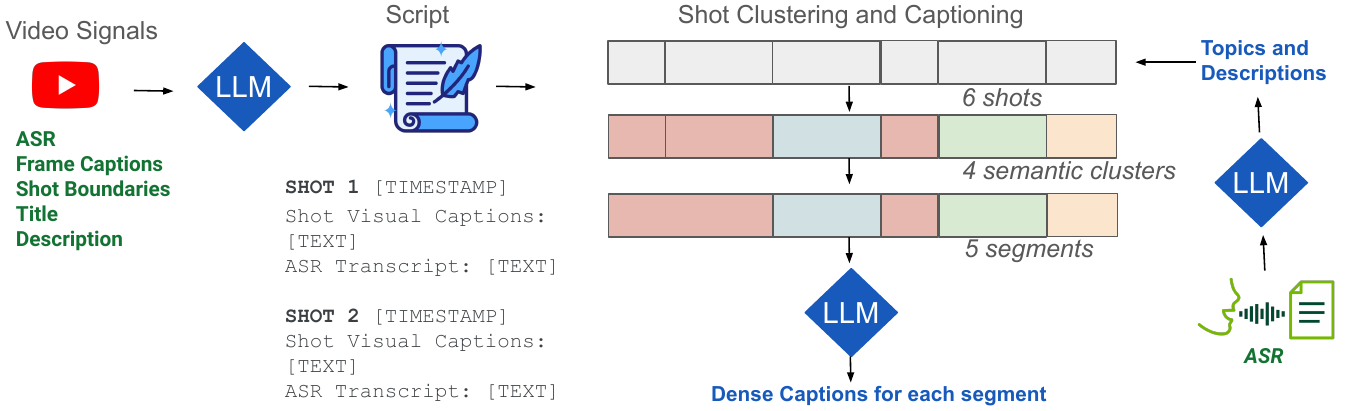}
    \caption{\textbf{Video Captioning:} We extract dense segment level captions automatically for each video. This is done by prompting an LLM using video signals (ASR, frame captions, shot boundaries and metadata) with various different steps and prompts.}
    \label{fig:videocap}
\end{figure}

A visual overview of the video captioning stage is provided in Fig. \ref{fig:videocap}. We describe the prompts for each stage below: \\
\noindent\textbf{Shot level captions:} \\
Using the shot boundaries the 1fps frame captions are summarized into shot level descriptions with the following prompt: \\
\begin{lmtt}
Summarize these sentences in dense short sentences:
[list of frame captions in the shot]
\end{lmtt}
\noindent\textbf{Topic and Description Pairs:} \\ 
If ASR exists, topic and description pairs are obtained from ASR using the following prompt: \\ 
\begin{lmtt}
**Task:**
Take a deep breath and give me the structural topics of the Youtube video below using the transcript.
Give up to 5 Topic and Description pairs using output format.
**Transcript:**
{transcript}
\end{lmtt}

\noindent\textbf{Shot Clustering:} \\
\begin{lmtt}
Take a deep breath and identify the sequential topic structure of this video using the "\{head\_topic\}" in Scenes.
A part of the video script is provided as a set of Scenes and in each scene, visual captions and transcript sentences are provided. The overall suggested structure from the transcript is provided as well.
Assign every scene in this part of the script to one topic structure.
For each scene, the visual captions should support and relate the topic. If the support or relation is not strong create a new topic and assign the scene to it.
Reevaluate the suggested structure from the transcript and make sure all scenes are assigned to the best associated topics.
Keep output length to be less than \{max\_output\_characters\} characters.

**Output Format:**
XML output where topic has the following children (description, topic\_scenes, story)
<topic>
<description>The description of the topic</description>
<topic\_scenes>Comma separated scene number(s) related to this topic<\/topic\_scenes>
<story>Summarized caption that describes what happens and what's shown for this topic in the scenes by combining visual caption and transcript sentences of the related scenes</story>
</topic>

**Suggested Structure:**
\{initial\_structure\_from\_ASR\_if\_exists\}

**Context:**
\{summary\_of\_title\_and\_description\}

**Video Script:**
\{video\_script\}
\end{lmtt}
\noindent\textbf{Segment Captions:} \\

Consecutive shots of the same topic are then merged as one segment. Shots of
the same topic that are not contiguous are treated as separate segments (see Fig. \ref{fig:videocap}).
We then generate dense captions for each segment using the prompt below: \\ 
\begin{lmtt}

**Task:**
You are the expert in video description writing. Use the information "Partial Script" to improve the "Initial Description" by adding the missing information either from visual or transcript. The video context is also given to help you interpret the script. Only add information that is in the "Partial Script".
Make the output concise and compact with less or the same length as Initial Description.
The updated video description is plain text.
Your answer should follow the output format.
Keep output length to be less than {max\_output\_characters} characters.

**Initial Description:***
{shared\_topic\_cluster\_caption}

**Output Format:**
XML format like below
<updated\_description>updated video description text</updated\_description>

**Partial Script:**
{doc\_segment}
\end{lmtt}
\noindent\textbf{Visual Support Caption} \\
To extract better visual description of the segment that will be used for QA generation in the next phase, an extra step is performed to get visual support for each segment. That visual support is stored separately in conjunction with the dense caption for the segment. For this purpose, the dense caption from the previous step is used alongside the shot level visual captions. The following LLM prompt is used to extract the visual support:
\\ 
\begin{lmtt}
**Task:**
I provide video scene information and your job is to summarize the exact elements from "Visual Captions" that directly support the "Scene Story" of the scene below.
The visuals of the scene is broken down to shots and each shot is described in a line of text in the Visual Captions.

**Scene Story:**
{dense\_caption\_for\_the\_segment}

**Visual Captions:**
{visual\_captions\_of\_the\_segment}

**Output Format:**
Plain text with at most 200 words summarizing the supporting visual elements.
\end{lmtt}

\subsubsection{Generating Questions and Answers} \label{sec:prompt-QAgen}
\begin{lmtt} 
I want you to act as a rigorous teacher in the "Long-term Video Understanding" class. Let's test your students' in-depth comprehension!

Understanding: I'll provide you with the following:

- Dense Captions: A detailed breakdown of the video, including key moments and timestamps. Analyze this carefully.

Your Task: Craft \{target\_number\} Challenging Short-Answer Questions

Requirement:

- Challenge: Demonstrate your ability to create challenging, insightful short-answer questions about the video. These shouldn't test simple recall only. Aim to probe understanding of relationships, motives, subtle details, and the implications of events within the video. \\
- Diversity: Design a variety of question types (more on this below). \\
- Specificity: Each question must be self-contained and laser-focused on a single concept or event from the video. Avoid compound or overly broad questions. \\
- Answers: Model the ideal answer format: Brief, accurate, and rooted directly in evidence from the video's content. \\
- Video-Centric: Stay true to what's explicitly shown or stated in the video. Avoid relying on outside knowledge or speculation. Design questions so the correct answer cannot be easily determined without carefully analyzing the video. \\
- Minimize Information Leakage: For question types like ranking or ordering, ensure that the order of candidates or options listed in the question doesn't inadvertently reveal the correct answer. Shuffle them to maintain neutrality. \\
- Content-First: Timestamps and section titles within the captions are there for guidance. Do not explicitly refer to those markers in your questions or answers. Focus on the events and elements themselves. \\
- Unambiguous: Ensure each question has a single, clearly defined correct answer. Avoid questions that are open to multiple interpretations (e.g., counting elements where viewers might disagree). \\
- Visual Elements: Questions focused on visual reasoning or visual narratives should emphasize the interpretation of the visuals. Keep the question minimal, letting the answer describe the specific visual elements in detail.

You want to test students’ capabilities of understanding the video, including but not limited to the following aspects:

Ability: Summarize and compare long parts of the video. \\
Ability: Compress information from the video rather than just listing the actions that happened in the video. \\
Ability: Identify the most important segments of the video. \\
Ability: Recognize and understand the visual elements in different parts of the video. \\
Ability: Understand the timeline of events and the plot in the video. \\
Ability: Count objects, actions and events. Focus on higher-level counting where the same instance does not occur in all/every frame and actions are sufficiently dissimilar. \\
Ability: Understand and reason about cause and effect in the video. \\
Ability: Understand the unspoken message that the audience may perceive after watching the video, which may require common sense knowledge to infer. \\
Ability: Understand the visual reasoning of why and how important visual content is shown in the video. \\
Ability: Understand the visual narrative of the video and the mood of the video and which visual elements do contribute to that. \\
Ability: Understand object states change over time, such as door opening and food being eaten.

Presentation

- QUESTION: Introduce each question as "QUESTION 1, 2, 3: (capability) full question".
- ANSWER: Follow the format "CORRECT ANSWER: correct answer".

Good example questions:
- Question (counting): How many ingredients are added to the bowl in total throughout the video?
  Correct Answer: 3.

- Question (goal reasoning): What is the purpose of the man standing in front of the whiteboard with a diagram on it?
  Correct Answer: To explain the features and capabilities of the vehicle.

- Question (cause and effect): How does the document help people to be happier?
  Correct Answer: It helps people to identify and focus on the things that make them happy, and to develop healthy habits.

- Question (timeline event): In what order are the following topics discussed in the video: history of pantomime, importance of pantomime, mime as a tool for communication, benefits of pantomime?
  Correct Answer: Mime as a tool for communication, history of pantomime, importance of pantomime, benefits of pantomime.

- Question (predictive): What happens after the man jumps up and down on the diving board?
  Correct Answer: He jumps into the pool.

- Question (summarization): What is the overall opinion of the reviewers about Hawaiian Shaka Burger?
  Correct Answer: The food is good, but the patties are frozen.

- Question (creator intent): What message does the video creators try to send to the viewers?
  Correct Answer: Nature is essential for human well-being.

- Question (visual-temporal): What color is the scarf that Jessica wears before she enters the restaurant?
  Correct Answer: Red.

- Question (visual narrative): How does John's overall facial expression contribute to the explanation of the financial situation that is described in the video?
  Correct Answer: He shows sad feelings and expression when he described the financial collapse of the company which adds to the sense of empathy that video describes.

- Question (visual reasoning): What was shown to support the effects of a high cholesterol diet in the video?
  Correct Answer: Video demonstrates how cholesterol gradually clogs blood vessels, using an animation to illustrate the cross-section of vessels and the buildup of plaque.

Bad example questions because it can be answered by common sense.
- Question (counting): How many players are there in a soccer team?
  Correct Answer: 11.

Bad example questions because it asks for trivial details.
- Question (counting): How many times the word 'hurricane' is said in the video?
  Correct Answer: 7.

Bad example questions because the summary of topics are subjective and ambiguous.
- Question (timeline event): List the sequence of topics Grace discusses in the video, starting with the earliest.
  Correct Answer: Getting ready for a photoshoot, attending a baseball game, showing off her new outfit, playing a Wayne's World board game, and discussing her upcoming week.

Dense Caption with Timestamps:
\{video\_inputs\_str\}

\end{lmtt} 

\subsubsection{Generating Decoys from Questions and Answers}

\begin{lmtt} 
Role: You are a rigorous teacher in a "Long-term Video Understanding" class. You will assist students in developing strong critical thinking skills. This requires creating sophisticated test questions to accompany video content.

Understanding: I will provide:

  - Dense Captions: A breakdown of the video, including structure, key events, and timestamps.
  - Target Questions \& Answers: A set of \{target\_number\} questions about the video, along with their correct answers.

Task: Generate High-Quality Multiple-Choice Questions

1.  Analyze: Carefully study the dense captions, questions, and correct answers. Familiarize yourself with the nuanced details of the video content.

2.  Decoy Design:  For each target question, generate \{decoy\_number\} incorrect answers (distractors). These distractors must be:\\
      - Challenging: Plausible to the point where students need deep content understanding and critical thinking to choose the correct answer. \\
      - Stylistic Match:  Mimic the style, tone, and complexity of the correct answer. \\
      - Similar Length:  Keep length close to that of the correct answer, preventing students from eliminating choices based on length differences. \\
      - Factually Relevant: Related to the video content, even if slightly incorrect due to a detail change, misinterpretation, or logical fallacy. \\
      - Reasonable:  Each decoy should be something that  could be true, making simple elimination impossible.

Specific Techniques for Distractor Creation

  - Subtle Tweaks: Alter a minor detail from the correct answer (e.g., change a time, location, or name).\\
  - Confusing Similarity:  Use a concept from elsewhere in the video that seems related but applies to a different context. \\
  - Misdirection:  Introduce a true statement related to the video's theme but not directly answering the question. \\
  - Order Shuffling: If the question involves the order of events, subtly rearrange the order within the distractors.

Presentation:

  - QUESTION: Repeat the provided question faithfully (e.g., "QUESTION 1 (Capability): ...") \\
  - CORRECT ANSWER:  Repeat the correct answer (e.g., "CORRECT ANSWER: ...") \\
  - WRONG ANSWERS: List each wrong answer on a separate line without using letters to label choices (e.g., "WRONG ANSWER 1: ...", "WRONG ANSWER 2: ...") \\

*GOOD* Example:
Question: What are the three main challenges that the college is taking on?
Correct Answer: Food scarcity, pollution, and disease.
Wrong Answer 1: Global warming, deforestation, and poverty.
Wrong Answer 2: Hunger, homelessness, and crime.
Wrong Answer 3: Obesity, malnutrition, and food insecurity.
Wrong Answer 4: Food waste, water shortages, and air pollution.

*BAD* examples where the decoys format is different from correct answer:
Question: What color is the shirt that the woman is wearing?
Correct Answer: Black.
Wrong Answer 1: The woman is wearing a white shirt.
Wrong Answer 2: The woman is wearing a blue shirt.
Wrong Answer 3: The woman is wearing a green shirt.
Wrong Answer 4: The woman is wearing a red shirt.

*BAD* examples because only the correct answer is in positive sentiment.
Question: What is the overall sentiment of the man in the video?
Correct Answer: He is overjoyed with his new gift.
Wrong Answer 1: He is upset his gift is not big enough.
Wrong Answer 2: He is sad about life in general.
Wrong Answer 3: He is upset the gift is not great.
Wrong Answer 4: He seems down and unhappy.

Dense Caption with Timestamps:
\{video\_inputs\_str\}

Question and Correct Answer:
\{question\_and\_answer\_str\}

\end{lmtt} 

\subsubsection{QAD Filtering}
The following prompt is used to filter out questions that can solve from QADs alone. 

\begin{lmtt} 
Instructions:

Carefully analyze the following question and options. Rank the options provided below, from the most likely correct answer to the least likely correct answer.
Please respond with "ANSWER" and "EXPLANATION".

Your response should be in the following format: \\
* ANSWER: [Letter of the ranking, split by greater than symbol. (e.g., "ANSWER: A > B > C > D > E")]. \\
* EXPLANATION: [Provide a brief explanation of your choice. Do not repeat the option.]

QUESTION:
\{question\_str\}

Options:
\{options\_str\}

Please provide your response below.
\end{lmtt}

\subsection{Human Rating and Correction of QADs}
\begin{figure} [h]
    \centering
\includegraphics[]{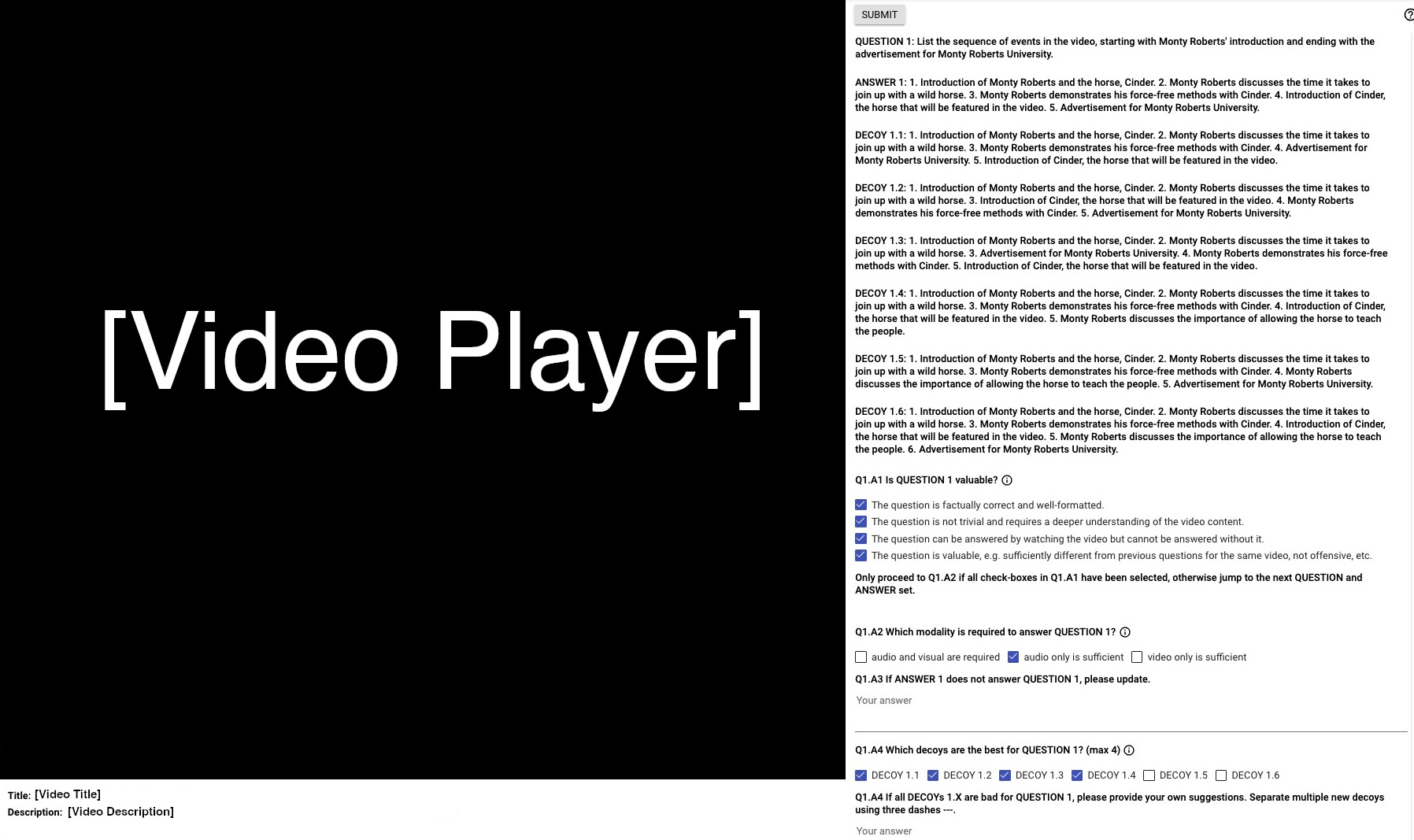}
    \caption{\textbf{Screenshot of rater UI.}}
    \label{fig:ui-screenshot}
\end{figure}

We provide a screenshot of the UI used by raters to annotate automatically generated QADs in Fig. \ref{fig:ui-screenshot}. Note that if any of the four options under the `Is the question valuable' field are not selected, then the question is discarded from the dataset. We made sure to train raters using training raters (with detailed decks and feedback rounds), as well as applying rater replication (we used 3 raters per question independently), and rater pipelining (having an experienced rater verify the answer from a previous rater) in order to correct hallucinations and other mistakes, and discard QADs that were inappropriate. Overall, of the total 11,030 QADs that we obtained automatically, 7,762 (~70\%) were discarded by raters. 

\subsection{Filtering Subsets}
\begin{figure*}
    \centering
    \begin{subfigure}[t]{0.5\textwidth}
    \includegraphics[width=\linewidth]{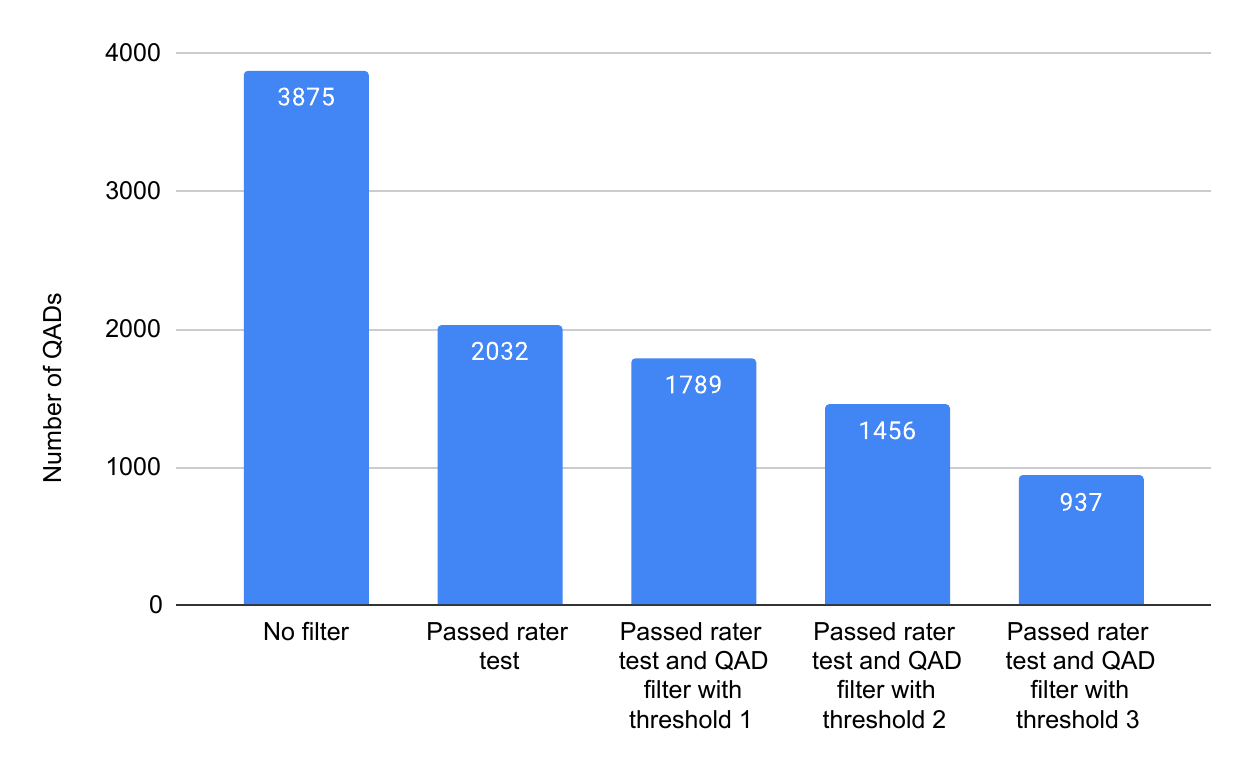}
    \caption{Effect of rater and QAD filters on dataset size}
    \label{fig:filter_thresholds_a}
    \end{subfigure}%
    ~
    \begin{subfigure}[t]{0.5\textwidth}
    \includegraphics[width=\linewidth]{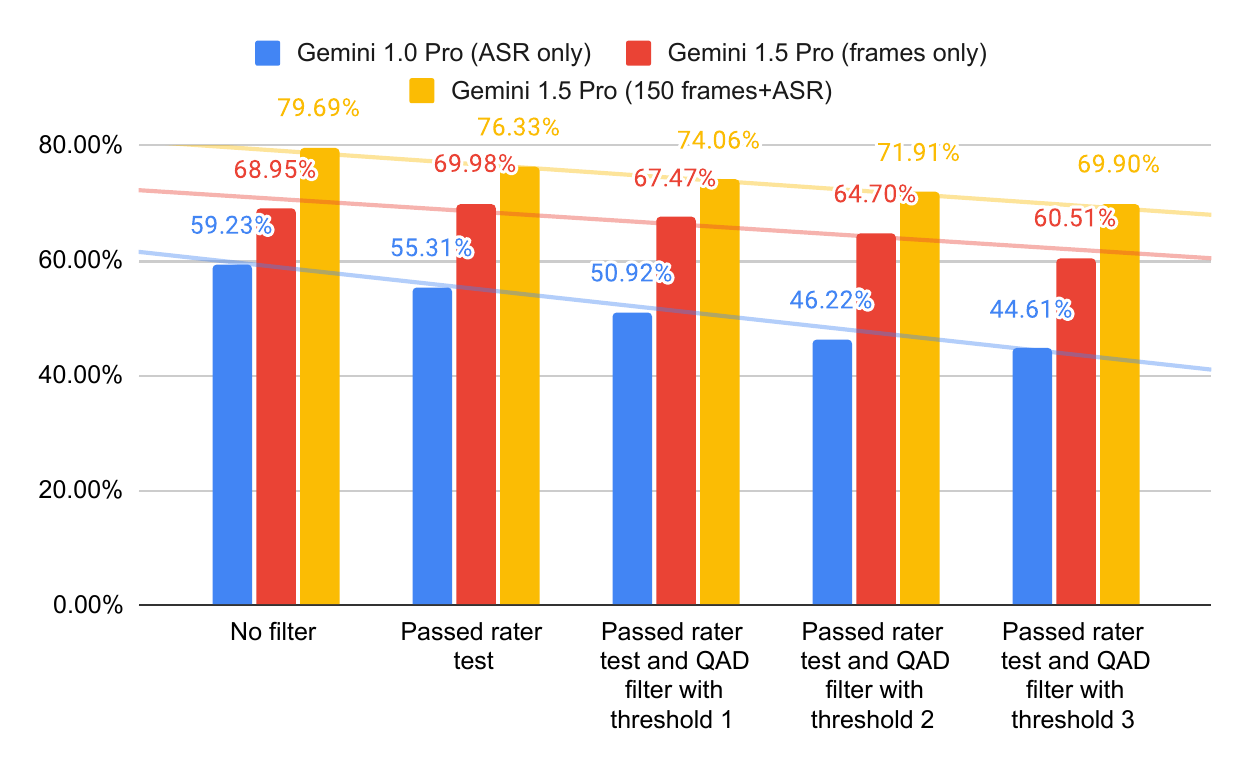}
    \caption{Effect of rater and QAD filters on accuracy scores}
    \label{fig:filter_thresholds_b}
    \end{subfigure}
    
    \begin{subfigure}[t]{0.5\textwidth}
    \includegraphics[width=\linewidth]{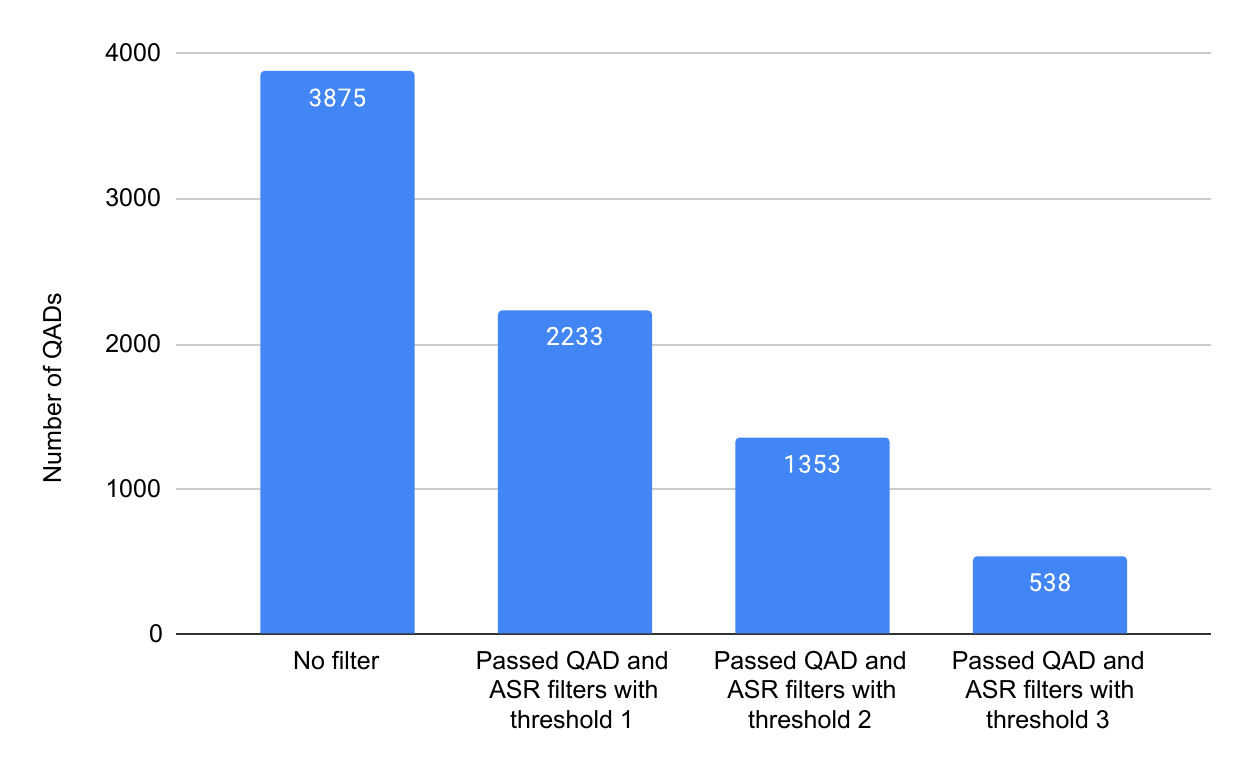}
    \caption{Effect of ASR and QAD filters on dataset size}
    \label{fig:filter_thresholds_c}
    \end{subfigure}%
    ~
    \begin{subfigure}[t]{0.5\textwidth}
    \includegraphics[width=\linewidth]{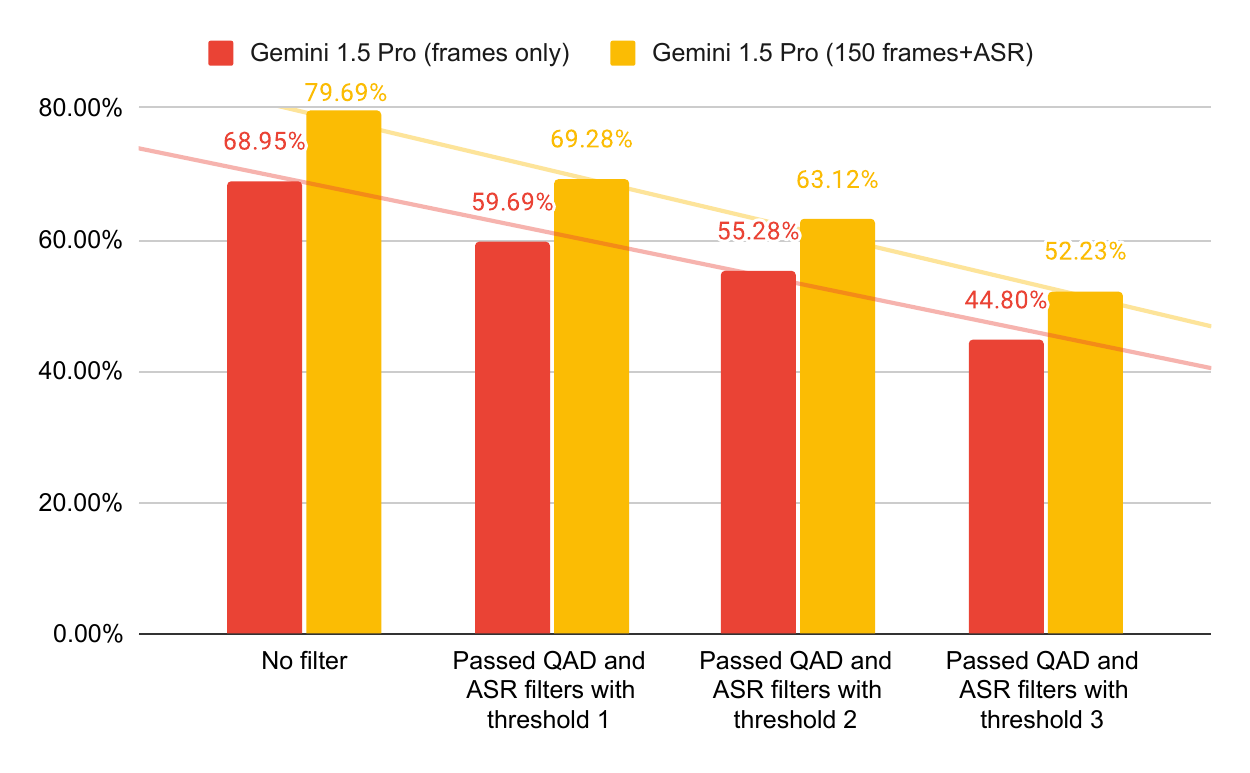}
    \caption{Effect of ASR and QAD filters on accuracy scores}
    \label{fig:filter_thresholds_d}
    \end{subfigure}
    \caption{Effect of filtering thresholds for the \textsc{Neptune-MMH} (top row) and \textsc{Neptune-MMA} (bottom row) subsets.}
    \label{fig:filter_thresholds}
\end{figure*}
Here we provide details for how we select the thresholds used to create the \textsc{Neptune-MMH} and \textsc{Neptune-MMA} subsets. For both subsets, we filtered \textsc{Neptune-Full} with the QAD filter described in Sec.~4.4. For \textsc{Neptune-MMA}, we additionally filtered out QADs that human raters marked as requiring only the audio modality and answer (see Sec.~4.5). We refer to this as the ``rater test''. For \textsc{Neptune-MMH}, we instead applied the ASR filter (Sec.~4.4). Both QAD and ASR filters were run by prompting an LLM (Gemini 1.0 Pro) three times, each time with a different random seed and then removing QADs that the LLM answered correctly at least X out of three times, where X is the threshold for the test.

Fig.~\ref{fig:filter_thresholds} shows how choosing different thresholds affects dataset size and accuracy scores. The top row shows the choices for the \textsc{Neptune-MMH} subset. Raters marked almost half of the questions as answerable from audio only, so the rater filter already cuts the dataset size in half. Successively applying the QAD filter with increasing thresholds reduces data size up until less than 25\%. We benchmark three models on the different subsets that have access to ASR only, vision only, or both vision and ASR, respectively. As expected, all three models show declining performance, with the ASR-only model showing the biggest losses. This suggests that all models were inferring the correct answer from the QAD only, which the filter successfully mitigates. The vision-only model gains slightly from removing QADs that fail the rater rest, which is expected as the test removes QADs that rely on audio, which the model does not have access to. However, like for the other models, its accuracy declines when adding the QAD test.

The bottom row of Fig.~\ref{fig:filter_thresholds} shows the choices for the \textsc{Neptune-MMA} subset where we use the ASR filter and the QAD filter with identical thresholds. This filter set has a stronger effect on the dataset size, reducing it to less than 15\% of its original size at the highest threshold. Because the ASR-only model was used for the ASR filter, we exclude it from the accuracy comparison. The vision-only and vision+ASR models both show declining accuracy with increasing thresholds. As expected, the accuracy of the vision+ASR model declines faster. The effect of this filter set on the accuracy is much stronger than that of the above filter set, suggesting that it increases the difficulty of the dataset more strongly. Even the vision-only model declines faster than above, suggesting that this filter set generally removes easier questions, even those that rely on vision only.

For both filtered sets, we opted to set the threshold to two, which in both cases significantly increases the dataset difficulty while still preserving enough QADs for statistically meaningful evaluation metrics. We noticed that when setting the threshold to three, there were less than five QADs left for some question types, preventing robust accuracy estimation for these tasks.

\subsection{Implementation Details for Benchmarks}
\subsubsection{Blind Baselines}
For the Gemini-1.5-pro baseline with text only the prompt used was: ``Carefully analyze the question and all available options then pick the most probable answer for this question''
\subsubsection{Video-LLaVA}
For Video-LLaVA the following prompt was used - 
"Pick a correct option to answer the question. Question: {question} Options: {options} ASSISTANT:".
\subsubsection{VideoLLama2}
During inference, we uniformly sampled 8 frames from each video. Each frame undergoes padding and resizing to a standardized dimension. The pre-processed frames are then fed into the image encoder. These steps are set as default in the inference script provided by videoLlama2. 

\textbf{QAD Prompt:} \_PROMPT\_TEMPLATE = """Pick a correct option number to answer the question. Question: \{question\} Options: \{options\}:"""

\textbf{OE Prompt:} Question: \{question\}

Output post processing: We eliminated extra characters and spaces using regex to get the final ID of the predicted option.

\subsubsection{MiniGPT4-Video} 

We set the 300 maximum number of output tokens to be 300 for the open-ended task and 10 for the multiple choice eval. 
The prompts are as follows: 

\_PROMPT\_TEMPLATE\_MCQ = """Question: select the correct option for this task: {question} Options: {options}. 
Output format: [OPTION]: [Reason]"""

\_PROMPT\_TEMPLATE\_OPEN\_ENDED = """Question: {question} Answer:"""

\subsubsection{MA-LMM} 
We set the 300 maximum number of output tokens to be 300 for the open-ended task and 300 for the multiple choice eval.
The prompts are as follows:

\_PROMPT\_TEMPLATE\_MCQ = """Question: select the best choice for this task: {question} Options: {options} Answer:"""

\_PROMPT\_TEMPLATE\_OPEN\_ENDED = """Question: {question} Answer:"""

\subsubsection{GPT-4o prompts}

\textbf{Open-ended evaluation with transcript}

\begin{lmtt}
You are an expert in video understanding and question answering. You can analyze a video given its image sequence and and transcript and answer questions based on them.

\{video\_frames\}

Video Transcript:
\{transcript\}

Answer the question using the image sequence. Do not describe the frames just answer the question.
Question: \{question\}
\end{lmtt}

\textbf{Open-ended evaluation without transcript}

\begin{lmtt}
You are an expert in video understanding and question answering. You can analyze a video given its image sequence and answer questions based on them.

\{video\_frames\}

Answer the question using the image sequence. Do not describe the frames just answer the question.
Question: \{question\}
\end{lmtt}

\textbf{Multiple-choice evaluation with transcript}

\begin{lmtt}
You are an expert in video understanding and question answering. You can analyze a video given its image sequence and and transcript and answer questions based on them.

\{video\_frames\}

Video Transcript:
\{transcript\}

Answer the question using the image sequence. Do not describe the frames just answer the question by identifying the choice.
Question: \{question\}
Choices: \{choices\}
Please identify the correct CHOICE and explain your reasoning concisely.
Output Format:
[CHOICE]: [REASON]
\end{lmtt}

\textbf{Multiple-choice evaluation without transcript}

\begin{lmtt}
You are an expert in video understanding and question answering. You can analyze a video as an image sequence and answer questions based on that.

\{video\_frames\}

Answer the question using the image sequence. Do not describe the frames just answer the question by identifying the choice.
Question: \{question\}
Choices: \{choices\}
Please identify the correct CHOICE and explain your reasoning concisely.
Output Format:
[CHOICE]: [REASON]
\end{lmtt}

\subsection{Compute Resources} 
The compute heavy part of the project was image frame captioning (as this involves reading high dimensional pixel data). The rest of the pipeline involves largely text-only LLMs and hence was less compute heavy. We estimate that the entire project in total took roughly 256 TPU v5e running over a period of 50 days.

\section{Additional details for \OEmetric}
\label{sec:appendix:gem}
\subsection{Creation of \OEmetric~equivalence dev set}
To create a development set that allows us to estimate the accuracy of different open-ended question answering metrics on Neptune, we sampled \new{97} question-answer pairs from the dataset and generated \new{3} candidate answers per question by prompting \new{VideoLLAVA~\citep{lin2023video}, Gemini-1.5-pro~\citep{reid2024gemini} and MA-LMM~\citep{he2024malmm} to write a free-form answer for each question without looking into the decoys or ground truth. We then manually annotated these responses between 0 and 1 by comparing it to the ground truth answer. We made sure that the annotators are blind to the model to avoid any bias. The resulting set has 292 equivalence pairs with an average score of 0.32, with 85 examples having score greater 0.5 and 206 examples with score less than 0.5}

\subsection{Benchmarking on the dev set}
In Table.~\ref{tab:gem-dev-results}, we evaluate several open-ended metrics on our dev set. The task of the metric is to classify whether the open-ended response and ground-truth answer are equivalent or not. We report F1-scores to balance false-positives and false-negatives. We evaluate both traditional rule-based metrics such as CIDEr and ROUGE-L, as well as established model-based metrics such as BEM\citep{bulian2022tomayto}. We also try using Gemini-1.5-pro \citep{reid2024gemini} as an LLM based equivalence metric (by prompting it to estimate equivalence).
First, we note that as expected, Gemini-1.5-pro correlates well with the human ground-truth annotation of the set, achieving a high F1-score of 72.5. However, given that Gemini is not open-source and proprietary, any change in the model can affect all the prior results in an external leader-board making it challenging as a metric. Traditional rule-based metrics perform much worse than Gemini-1.5-pro on this dev set as they are n-gram based and struggle to handle the diversity of domains and styles in the open-ended responses. The BERT model based BEM metric \citep{bulian2022tomayto} performs similarly, achieving an F1-score of 61.5. 

\new{Next, we evaluate lightweight open-source language models Gemma-v1-2B \citep{team2024gemma},  Gemma-v1-7B \citep{team2024gemma} and Gemma-v2-9B \citep{gemmateam2024gemma2} in a zero-shot setting and find that performance improves with model size, with Gemma-v2-9B bridging the gap well between traditional metrics and the Gemini-1.5-pro based metric. Finally, we fine-tune Gemma-v2-9B on the open-source BEM answer equivalence dataset \citep{bulian2022tomayto}, and find that Gemma-v2-9B finetuned on the BEM dataset performs the best on our dev-set. We name this metric \textit{\OEmetric}. }

\subsection{Implementation Details}
We use instruction-tuned variants of the Gemma models (gemma-v1-it-2b, gemma-v1-it-7b and gemma-v2-it-9b) for our experiments. To develop a prompt, we experiment with several variations in a zero-shot setting and measure the performance on the dev-set. Our final prompt is shown below. To ensure responses occur in a standard format, we simply measure the softmax-probability over "TRUE" response indicating the statements are equivalent and "FALSE" response indicating the statements are not equivalent. For each model, the threshold over probability is chosen to maximize the F-1 score on dev set. To finetune Gemma models on BEM dataset, we tokenize the same prompt as used in the zero-shot setting and train it using prefix-LM tuning for $10000$ iterations using a learning rate of $1e-6$. For evaluation, we truncate the open-ended responses to $100$ words, use a decode cache size of $1024$ and threshold the softmax probability of the LM using the chosen threshold from dev-set. 

\begin{verbatim}
<start_of_turn>user
Answer Equivalence Instructions:

Carefully consider the following question and answers.
You will be shown a "gold-standard" answer from a human annotator,
referred to as the "Reference Answer" and a "Candidate Answer".
Your task is to determine whether the two answers are semantically
equivalent.

In general, a candidate answer is a good answer in place of the "gold"
reference if both the following are satisfied:
  1. The candidate contains at least the same (or more) relevant information
     as the reference,taking into account the question; in particular it
     does not omit any relevant information present in the reference.
  2. The candidate contains neither misleading or excessive superfluous
     information not present in the reference, taking into account the
     question.

Your response should be one word, "TRUE" or "FALSE", in the following
format:
ANSWERS_ARE_EQUIVALENT: [TRUE or FALSE]

Question:
"{}"

Candidate Answer:
"{}"

Reference Answer:
"{}"

Please provide your response below.
<end_of_turn>
<start_of_turn>model
ANSWERS_ARE_EQUIVALENT: 
\end{verbatim}

\subsection{Qualitative Examples for Metric}
\label{sec:appendix:gem-examples}
\new{Below, we provide some qualitative examples of our metric applied to the predictions provided by the baseline models  Video LLaMA2\citep{cheng2024videollama2},  Gemini-1.5-flash~\citep{reid2024gemini} and Gemini.1.5-pro~\citep{reid2024gemini}.} We find that Gemini-1.5-flash responses are qualitatively slightly better than Gemini-1.5-pro responses as they contain less hallucinations. Furthermore, we find that with a single frame as input, Gemini-1.5-pro usually responds with a variation of "This information is not present is the provided frame" instead of making an educated guess. This is unlike the case of multiple choice questions where the model commits to one of the choices. These aspects are reflected in the results presented in Table.\ref{tab:results_fixed} and Table.\ref{tab:mod_ablations}. We also find that the metric fails often in the questions about listing items / events in order. In such cases, the model is biased towards accepting the response (false positives) even if the order is incorrect provided the items / events are present in the candidate response. 

\begin{verbatim}
Question: Compare the Pampered Chef Veggie Strip Maker to a food
    processor in terms of their effectiveness in making cauliflower
    rice.
Answer: The Veggie Strip Maker is more effective than a food
    processor in making cauliflower rice because it prevents the
    cauliflower from becoming mushy.
Video LLaMA2: Both the Pampered Chef Veggie Strip Maker and a food
    processor can be effective in making cauliflower rice, but they
    differ in their methods of operation and the resulting texture of
    the rice.    The Pampered Chef Veggie Strip Maker is a kitchen
    appliance designed specifically for making vegetable strips,
    including cauliflower rice. It uses a grating mechanism to create
    rice-like pieces by shredding the cauliflower. This method can
    produce a relatively uniform and rice-like texture.    On the other
    hand, a food processor is a versatile kitchen appliance that can
    also be used to make ...
GEM Equivalence: False
Gemini-1.5-flash: The Pampered Chef Veggie Strip Maker makes
    cauliflower rice that is less mushy than a food processor.
GEM Equivalence: True
Gemini-1.5-pro: The Pampered Chef Veggie Strip Maker is more
    effective than a food processor because it doesn't make the
    cauliflower mushy.
GEM Equivalence: True


Question: How does the use of behavioral training help in the
    process of taking foot x-rays on elephants?
Answer: Behavioral training encourages elephants to willingly
    participate in their healthcare, making the process of taking foot
    x-rays less stressful for both the elephants and the caretakers.
Video LLaMA2: The use of behavioral training plays a crucial role in
    the process of taking foot x-rays on elephants. By training the
    elephants to follow commands and stand on a platform, it allows the
    caretakers to safely and effectively capture the x-rays of the
    elephants' feet. This training helps to ensure that the elephants
    remain still and cooperative during the x-ray process, minimizing
    any potential harm or discomfort. Additionally, the behavioral
    training helps to build a positive relationship between the
    elephants and their caretakers, fostering trust and cooperation.
    Overall, the use of behavioral training enhances the accuracy and
    efficiency of the
GEM Equivalence: True
Gemini-1.5-flash: Behavioral training allows the elephants to
    willingly participate in the training sessions and participate in
    their own health care.
GEM Equivalence: True
Gemini-1.5-pro: Behavioral training allows the elephants to
    willingly participate in the x-ray process by presenting their feet
    for imaging.
GEM Equivalence: True

Question: What are the key features of the Adam Purvis Progeny
    knife's blade?
Answer: The Adam Purvis Progeny knife's blade is made of 3.4-inch
    M390 steel. The blade shape is useful, simple, and highly effective,
    the edge is very sharp, featuring a stonewash finish on the primary
    bevels and a horizontal satin on the flats. It is exceptionally
    sharp and has a 150 thousandths stock thickness, making it suitable
    for everyday carry tasks.
Video LLaMA2: The Adam Purvis Progeny knife's blade is made of S35VN
    steel, which is known for its high strength and excellent edge
    retention. The blade has a flat ground and a hollow grind, which
    allows for a thin and sharp edge. Additionally, the blade has a
    black and silver color scheme, giving it a sleek and modern
    appearance.
GEM Equivalence: False
Gemini-1.5-flash: The Adam Purvis Progeny knife's blade is made of
    M390 steel, has a modified sheepsfoot style, and is ground very
    thinly for a sharp edge.
GEM Equivalence: False
Gemini-1.5-pro: The Adam Purvis Progeny knife features a 3.4-inch
    M390 steel blade with a modified sheepsfoot shape, thin grind, sharp
    edge, and a stonewashed finish on the bevels with a satin finish on
    the flats.
GEM Equivalence: True


Question: How does interacting with civilians in nearby villages
    enhance cadets' leadership skills?
Answer: Engaging with civilians in nearby villages allows cadets to
    develop respectful communication and relationship-building skills,
    which are essential for effective leadership in various situations.
Video LLaMA2: Interacting with civilians in nearby villages enhances
    cadets' leadership skills by providing them with real-world
    experiences and opportunities to apply their knowledge and training
    in a practical setting. By engaging with the local community, cadets
    learn to communicate effectively, build trust, and foster positive
    relationships with people from diverse backgrounds. This hands-on
    experience helps them develop their interpersonal skills, cultural
    awareness, and problem-solving abilities, all of which are crucial
    for successful leadership in various situations. Additionally,
    interacting with civilians allows cadets to gain a deeper
    understanding of the challenges and needs of the community, which
    can inform their decision-making and ...
GEM Equivalence: True
Gemini-1.5-flash: Interacting with civilians in nearby villages
    teaches cadets how to speak to people respectfully and build trust.
GEM Equivalence: True
Gemini-1.5-pro: It teaches them how to speak to people.
GEM Equivalence: False


Question: List the sequence of topics discussed in the video,
    starting with the earliest.
Answer: 1. Introduction of the GAN356 AIR cube and its unique
    features. 2. Discussion of the GAN356 AIR's weight and GES (Gan
    Elasticity System) springs. 3. Lubrication, tensioning, and personal
    experience with the GAN356 AIR's performance. 4. Assessment of the
    GAN356 AIR's performance, including its smoothness, speed, corner-
    cutting, and tendency to lock up. 5. Overall evaluation of the
    GAN356 AIR, including its strengths and weaknesses, and
    recommendations for potential buyers.
Video LLaMA2: 1. Introduction to the video and the Rubik's Cube.  2.
    Explanation of the different types of Genuine Handcubes.  3.
    Demonstration of how to solve the Rubik's Cube, including the
    various algorithms and techniques used.  4. Showcasing of the
    different colors and types of Rubik's Cubes available.  5.
    Conclusion and closing credits.
GEM Equivalence: False
Gemini-1.5-flash: The video starts with an introduction of the GAN
    356 Air cube, then discusses its features, including the GES system
    and its light weight. The reviewer then shares their opinion on the
    different GES springs and their effect on the cube's feel. The video
    continues with a discussion on the cube's performance, including its
    smoothness, speed, and tendency to lock up. Finally, the reviewer
    concludes with a recommendation for viewers who might be interested
    in purchasing the cube.
GEM Equivalence: True
Gemini-1.5-pro: The video discusses the main features of the Gan 356
    Air, the different types of springs, the weight and price of the
    cube, the reviewer's experience with the cube, and the reviewer's
    overall opinion of the cube.
GEM Equivalence: False



Question: In what order are the following events mentioned in the
    video: introduction of the wallet, features of the wallet, and care
    instructions?
Answer: Introduction of the wallet, features of the wallet, care
    instructions.
Video LLaMA2: The video begins with the introduction of the wallet,
    followed by a description of its features, and concludes with
    instructions on how to care for the wallet.
GEM Equivalence: True
Gemini-1.5-flash: The speaker introduces the wallet, then describes
    its features, and finally mentions the care instructions.
GEM Equivalence: True
Gemini-1.5-pro: Introduction of the wallet, care instructions,
    features of the wallet.
GEM Equivalence: False



\end{verbatim}

\section{Model-Based Temporal Certificate}
\label{sec:appendix-temporal-certificate}
The idea of a temporal certificate was introduced by EgoSchema~\citep{mangalam2023egoschema} as a way of capturing the intrinsic temporal understanding length for \textit{long video} QA datasets. It is defined as `the length of the video a human verifier needs to observe to be convinced of
the veracity of the marked annotation'. While the authors used it to uncover flaws in existing long video QA datasets, as well as to provide a difficulty measure independent of video length, we find that is has the following drawbacks: (i) it does not take into account the \textit{length of time} or the \textit{effort} taken by the annotator themselves, to find the correct time span in videos;  (ii) it requires manual annotation from expert annotators to measure; and finally (iii) is subjective.

\begin{figure*}[h]
\centering
  \includegraphics[width=0.5\linewidth]{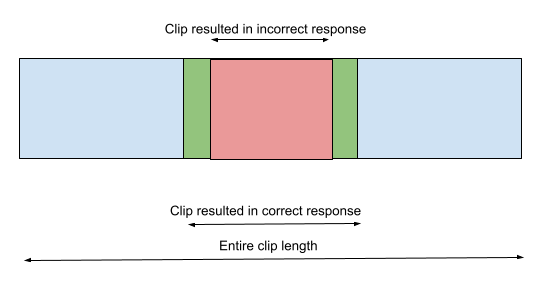}
  \caption{\textbf{Model-based Temporal Certificate:} Illustration of video clip querying for the model-based temporal certificate experiment. The red clip is the clip length that resulted in an incorrect response. As we increased the clip length wider, and the model correctly answered the question, we logged the frame count for incorrect response and correct response, and stopped querying. Besides clip length, we vary the fps of the query clip.
  }
  \label{fig:temporal-certificate-experiment}
\end{figure*}

As an attempt to mitigate these issues, we introduce a slightly modified version of the temporal certificate, which is \textit{Model-Based}. We calculate this certificate using 129 samples from Neptune and EgoSchema, respectively. For this experiment we used Gemini 1.5 Pro, with one ``driver'' model run to answer the question and two other model runs with different random seeds to verify if the answer was not correct by random chance. Along with the question and options, we provided video clips of various lengths from the center of the video, and at various fps, as shown in Fig.~\ref{fig:temporal-certificate-experiment}.

Since this experiment queried a set of frames over various clip lengths, we defined it as the “needle in haystack” problem. Here, the needle is defined as a frame or set of frames needed to answer the question correctly, matching a human's ground truth response, while the haystack is a set of frames which need to be watched to find the needle frames. Iteratively, we increase the video length and fps for the query until the model achieves the correct response.

\begin{figure*}[t]
\centering
    \begin{subfigure}[t]{0.5\textwidth}
    \includegraphics[width=\linewidth]{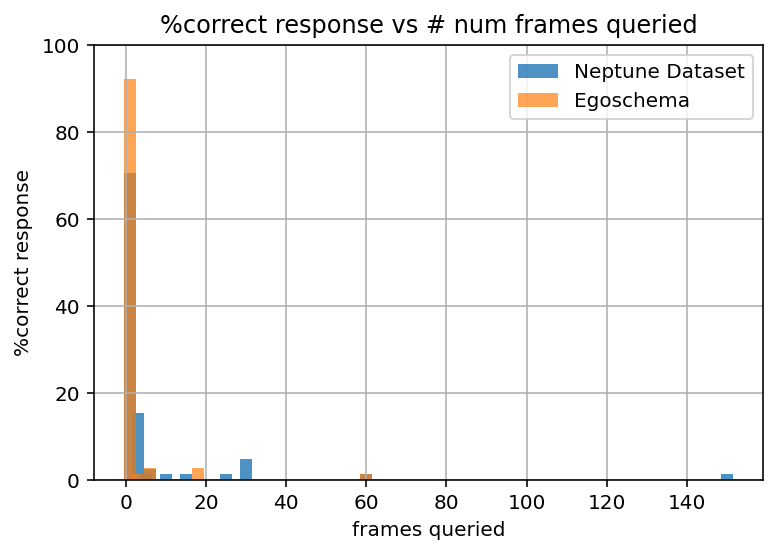}
    \label{fig:num_frames_a}
    \end{subfigure}%
    ~
    \begin{subfigure}[t]{0.5\textwidth}
    \includegraphics[width=\linewidth]{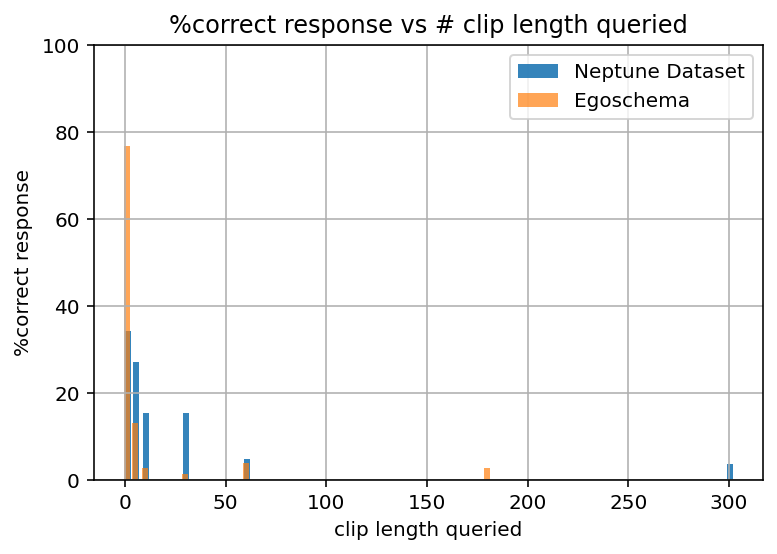}
    \label{fig:num_frames_b}
    \end{subfigure}
  \caption{\textbf{Frame level temporal certificate:} We compared our dataset sample with EgoSchema to evaluate the number of frames needed by model to answer questions correctly. The figures above show the distribution of the minimum number of frames required to achieve the correct response.
  }
  \label{fig:num-frames}
\end{figure*}

As shown in Fig.~\ref{fig:num-frames}, 
we find that the model needs more frames to answer the question correctly for the Neptune dataset as compared to EgoSchema. This resulted in a mean of 5.39 as certificate frames for Neptune which is 3.37 times the mean certificate frame number of 1.6 for EgoSchema. On the clip length level this translated to a mean of 21.22s of clip needed to respond correctly on the Neptune dataset, whereas for EgoSchema the mean was 9.07s. The model-based certificate lengths turn out to be much smaller than the certificate lengths reported by EgoSchema, where humans needed close to 100s to answer the questions for EgoSchema.

In addition, we define the \emph{effort score} as the fraction of the maximum number of frames needed to be watched before answering the question correctly, as defined in Equation \ref{eq:eff-score}.
An effort score closer to 0 suggests that the needle isn't very small compared to the haystack, i.e. most of the frames contain the answer to the question; while a high effort score means a high percentage of haystack frames needs to be included before we cover all frames required to answer correctly.

\begin{equation}
  \textsc{effort\ score} = \frac{\textsc{max\ number\ of\ frames\ resulting\ in\ an\ incorrect\ response}}{\textsc{min\ number\ of\ frames\ resulting\ in\ a\ correct\ response}}
  \label{eq:eff-score}
\end{equation}

For Neptune, the mean effort score was 0.47, whereas for EgoSchema, it was 0.19. This suggests that Neptune requires 2.47 times the effort compared to EgoSchema according to the definition above, which closely corroborates the above results for the mean clip lengths needed to solve the questions from the respective datasets.

\section{Examples of caption quality}
\begin{figure*}[t]
  \includegraphics[width=\linewidth]{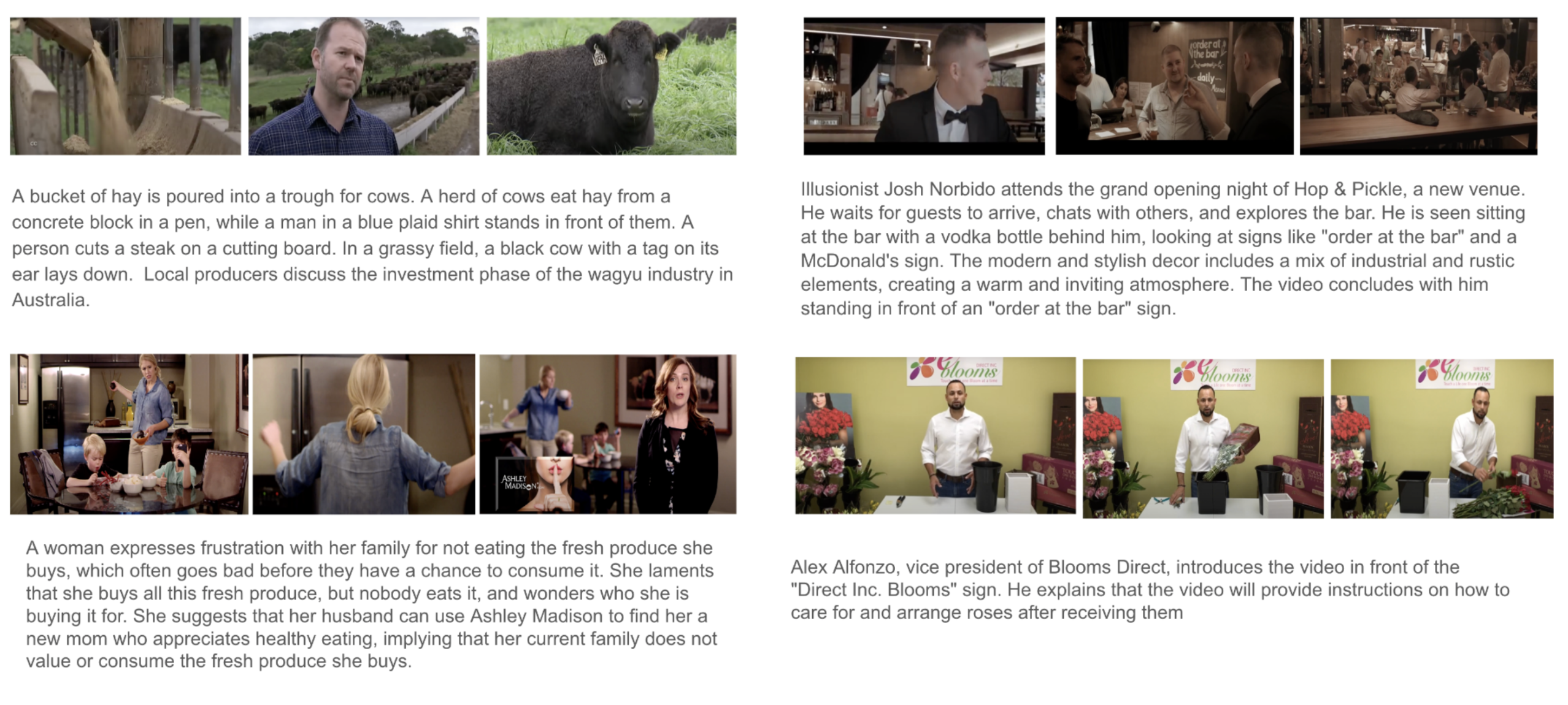}
  \caption{\textbf{Qualitative results of automatic caption generation.} Best viewed zoomed in. Note how the captions contain plenty of visual details, can contain numerous different events (top left), can mention mood and atmosphere (top right), use details from the ASR, and can even mention high level feelings and emotions (bottom left). Bottom right shows a failure case, where the caption is accurate, but too simple and high level and does not cover the fine-grained actions that the man takes.}
  \label{fig:caption_qual}
\end{figure*}
We show examples of captions generated by our automatic pipeline in Fig.~\ref{fig:caption_qual}.

\section{Societal Impact}
\label{sec:societal_impact}
Our data may match the distribution of videos and text on the internet. As such, it will mirror known biases on that source of data. For at least this reason, this data set should not be used for training models and is only intended for academic evaluation purposes. 
To create the dataset, we run large Gemini models, which has a negative externality of energy usage and carbon emissions.
For benchmarking, we use existing models. These models are likely to inherit the biases of the data distribution and the pre-trained weights used in their original training.

\end{document}